\documentclass[10pt,twocolumn,letterpaper]{article}

\usepackage{iccv}
\usepackage{times}
\usepackage{epsfig}
\usepackage{graphicx}
\usepackage{amsmath}
\usepackage{amssymb}
\usepackage[misc]{ifsym}
\usepackage{algorithm}
\usepackage{algorithmicx}
\usepackage{algpseudocode}
\usepackage{spverbatim}
\usepackage{etoolbox}
\makeatletter
\preto{\@verbatim}{\topsep=1pt \partopsep=1pt}
\makeatother

\usepackage{booktabs}
\usepackage[x11names,dvipsnames,table]{xcolor} 
\usepackage{multirow}
\usepackage{array}
\newcolumntype{L}[1]{>{\raggedright\let\newline\\\arraybackslash\hspace{0pt}}m{#1}}
\newcolumntype{C}[1]{>{\centering\let\newline\\\arraybackslash\hspace{0pt}}m{#1}}
\newcolumntype{R}[1]{>{\raggedleft\let\newline\\\arraybackslash\hspace{0pt}}m{#1}}

\newcommand{\figurewidth}{8.0cm}
\newcommand{\scatterwidth}{4.0cm}
\newcommand{\scatterwidthA}{4.0cm}
\newcommand{\scatterwidthB}{4.0cm}
\newcommand{\subfigurehkern}{0.3cm}
\newcommand{\subfigurevkern}{0.2cm}
\newcommand{\colwidth}{1.0cm}
\newcommand{\colwidthA}{1.0cm}
\newcommand{\colwidthB}{1.0cm}

\usepackage[breaklinks=true,bookmarks=false]{hyperref}

\iccvfinalcopy 

\def\iccvPaperID{****} 

\ificcvfinal\fi
\begin{document}

\title{SORT: Second-Order Response Transform for Visual Recognition}

\author{Yan Wang\textsuperscript{1}, Lingxi Xie\textsuperscript{2}$^{(\textrm{\Letter})}$,
Chenxi Liu\textsuperscript{2}, Siyuan Qiao\textsuperscript{2} \\
Ya Zhang\textsuperscript{1}$^{(\textrm{\Letter})}$, Wenjun Zhang\textsuperscript{1},
Qi Tian\textsuperscript{3}, Alan Yuille\textsuperscript{2} \\
\textsuperscript{1}{Cooperative Medianet Innovation Center, Shanghai Jiao Tong University, Shanghai, China} \\
\textsuperscript{2}{Department of Computer Science, The Johns Hopkins University, Baltimore, MD, USA} \\
\textsuperscript{3}{Department of Computer Science, The University of Texas at San Antonio, San Antonio, TX, USA} \\
{\tt\small tiffany940107@gmail.com}\quad{\tt\small 198808xc@gmail.com}\quad{\tt\small \{cxliu,siyuan.qiao\}@jhu.edu} \\
{\tt\small \{ya\_zhang,zhangwenjun\}@sjtu.edu.cn}\quad{\tt\small qitian@cs.utsa.edu}\quad{\tt\small alan.l.yuille@gmail.com} \\
}

\maketitle
\thispagestyle{empty}

\begin{abstract}
In this paper, we reveal the importance and benefits of introducing second-order operations into deep neural networks.
We propose a novel approach named Second-Order Response Transform (SORT),
which appends element-wise product transform to the linear sum of a two-branch network module.
A direct advantage of SORT is to facilitate cross-branch response propagation,
so that each branch can update its weights based on the current status of the other branch.
Moreover, SORT augments the family of transform operations and increases the nonlinearity of the network,
making it possible to learn flexible functions to fit the complicated distribution of feature space.
SORT can be applied to a wide range of network architectures,
including a branched variant of a chain-styled network and a residual network, with very light-weighted modifications.
We observe consistent accuracy gain on both small (CIFAR10, CIFAR100 and SVHN) and big (ILSVRC2012) datasets.
In addition, SORT is very efficient, as the extra computation overhead is less than $5\%$.
\end{abstract}

\section{Introduction}
\label{Introduction}

Deep neural networks~\cite{Krizhevsky_2012_ImageNet}\cite{Simonyan_2015_Very}\cite{Szegedy_2015_Going}\cite{He_2016_Deep}
have become the state-of-the-art systems for visual recognition.
Supported by large-scale labeled datasets such as {\bf ImageNet}~\cite{Deng_2009_ImageNet}
and powerful computational resources like modern GPUs,
it is possible to train a hierarchical structure to capture different levels of visual patterns.
Deep networks are also capable of generating transferrable features
for different vision tasks such as image classification~\cite{Donahue_2014_DeCAF} and instance retrieval~\cite{Razavian_2014_CNN},
or fine-tuned to deal with a wide range of challenges,
including object detection~\cite{Girshick_2014_Rich}\cite{Ren_2015_Faster},
semantic segmentation~\cite{Long_2015_Fully}\cite{Chen_2015_Semantic},
boundary detection~\cite{Shen_2015_Deepcontour}\cite{Xie_2015_Holistically}, {\em etc}.

The past years have witnessed an evolution in designing efficient network architectures,
in which the chain-styled modules have been extended to
multi-path modules~\cite{Szegedy_2015_Going} or residual modules~\cite{He_2016_Deep}.
Meanwhile, highway inter-layer connections are verified helpful in training very deep networks~\cite{Srivastava_2015_Highway}.
In the previous literatures, these connections are fused in a linear manner,
{\em i.e.}, the neural responses of two branches are element-wise summed up as the output.
This limits the ability of a deep network to fit the complicated distribution of feature space,
as nonlinearity forms the main contribution to the network capacity~\cite{Jarrett_2009_What}.
This motivates us to consider higher-order transform operations.

\renewcommand{\figurewidth}{8.0cm}
\begin{figure}
\begin{center}
    \includegraphics[width=\figurewidth]{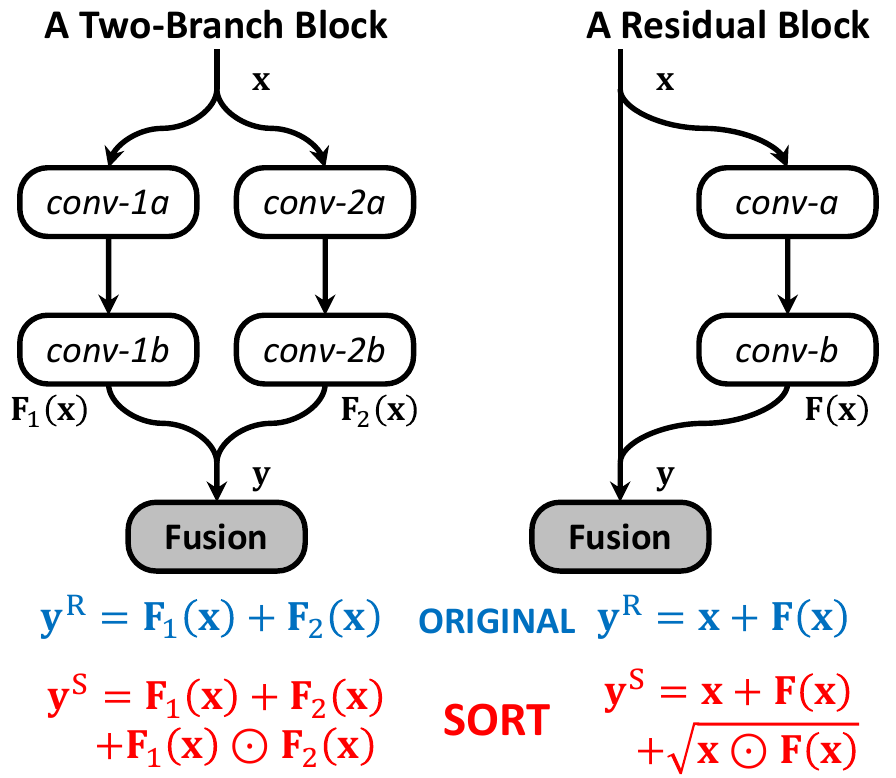}
\end{center}
\caption{
    Two types of modules and the corresponding SORT operations.
    Left: in a two-branch convolutional block, the two-way outputs,
    $\mathbf{F}_1\!\left(\mathbf{x}\right)$ and $\mathbf{F}_2\!\left(\mathbf{x}\right)$,
    are combined with a second-order transform $\mathbf{F}_1\!\left(\mathbf{x}\right)+\mathbf{F}_2\!\left(\mathbf{x}\right)+
    \mathbf{F}_1\!\left(\mathbf{x}\right)\odot\mathbf{F}_2\!\left(\mathbf{x}\right)$.
    Right: in a residual-learning building block~\cite{He_2016_Deep},
    we can also modify the fusion stage from $\mathbf{x}+\mathbf{F}\!\left(\mathbf{x}\right)$
    to $\mathbf{x}+\mathbf{F}\!\left(\mathbf{x}\right)+\sqrt{\mathbf{x}\odot\mathbf{F}\!\left(\mathbf{x}\right)}$.
    Here, $\odot$ denotes element-wise product, and $\sqrt{\cdot}$ denotes element-wise square-root.
}
\label{Fig:SORT}
\end{figure}

In this paper, we propose Second-Order Response Transform ({\bf SORT}),
an efficient approach that applies to a wide range of visual recognition tasks.
The core idea of SORT is to append a dyadic second-order operation, say element-wise product,
to the original linear sum of two-branch vectors.
This modification, as shown in Figure~\ref{Fig:SORT}, brings two-fold benefits.
First, SORT facilitates {\em cross-branch} information propagation,
which rewards consistent responses in forward-propagation,
and enables each branch to update its weights based on the current status of the other branch in back-propagation.
Second, the nonlinearity of the module becomes stronger, which allows the network to fit more complicated feature distribution.
In addition, adding such operations is very cheap,
as it requires less than $5\%$ extra time, and no extra memory consumptions.
We apply SORT to both deep chain-styled networks and deep residual networks,
and verify consistent accuracy gain over some popular visual recognition datasets,
including {\bf CIFAR10}, {\bf CIFAR100}, {\bf SVHN} and {\bf ILSVRC2012}.
SORT also generates more effective deep features to boost the transfer learning performance.

The remainder of this paper is organized as follows.
Section~\ref{RelatedWork} briefly reviews related work,
and Section~\ref{Approach} illustrates the SORT algorithm and some analyses.
Experiments are shown in Section~\ref{Experiments}, and conclusions are drawn in Section~\ref{Conclusions}.

\section{Related Work}
\label{RelatedWork}

\subsection{Convolutional Neural Networks}
\label{RelatedWork:CNN}

The Convolutional Neural Network (CNN) is a hierarchical model for visual recognition.
It is based on the observation that a deep network with enough neurons is able to fit any complicated data distribution.
In past years, neural networks were shown effective for simple recognition tasks~\cite{LeCun_1990_Handwritten}.
More recently, the availability of large-scale training data ({\em e.g.}, ImageNet~\cite{Deng_2009_ImageNet}) and powerful GPUs
make it possible to train deep architectures~\cite{Krizhevsky_2012_ImageNet} which significantly outperform
the conventional Bag-of-Visual-Words~\cite{Lazebnik_2006_Beyond}\cite{Wang_2010_Locality}\cite{Perronnin_2010_Improving}
and deformable part models~\cite{Felzenszwalb_2010_Object}.
A CNN is composed of several stacked layers.
In each of them, responses from the previous layer are convolved with a filter bank and activated by a differentiable non-linearity.
Hence, a CNN can be considered as a composite function,
which is trained by back-propagating error signals defined by the difference between supervision and prediction at the top layer.
Recently, efficient methods were proposed to help CNNs converge faster and prevent over-fitting,
such as ReLU activation~\cite{Nair_2010_Rectified}, Dropout~\cite{Srivastava_2014_Improving},
batch normalization~\cite{Ioffe_2015_Batch} and varying network depth in training~\cite{Huang_2016_Deep}.
It is believed that deeper networks have stronger ability
of visual recognition~\cite{Simonyan_2015_Very}\cite{Szegedy_2015_Going}\cite{He_2016_Deep},
but at the same time, deeper networks are often more difficult to be trained efficiently~\cite{Srivastava_2015_Training}.

An intriguing property of the CNN lies in its transfer ability.
The intermediate responses of CNNs can be used as effective image descriptors~\cite{Donahue_2014_DeCAF},
and widely applied to various types of vision applications,
including image classification~\cite{Jia_2014_CAFFE}\cite{Xie_2016_InterActive}
and instance retrieval~\cite{Razavian_2014_CNN}\cite{Xie_2015_Image}.
Also, deep networks pre-trained on a large dataset can be fine-tuned to deal with other tasks,
including object detection~\cite{Girshick_2014_Rich}\cite{Ren_2015_Faster},
semantic segmentation~\cite{Chen_2015_Semantic}, boundary detection~\cite{Xie_2015_Holistically}, {\em etc}.

\subsection{Multi-Branch Network Connections}
\label{RelatedWork:MultiBranch}

Beyond the conventional chain-styled networks~\cite{Simonyan_2015_Very},
it is observed that adding some sideway connections can increase the representation ability of the network.
Typical examples include the inception module~\cite{Szegedy_2015_Going},
in which neural response generated by different kernels are concatenated to convey multi-scale visual information.
Meanwhile, the benefit of identity mapping~\cite{He_2016_Identity} motivates researchers
to explore networks with residual connections~\cite{He_2016_Deep}\cite{Zagoruyko_2016_Wide}\cite{Huang_2017_Densely}.
These efforts can be explained as the pursuit of building highway connections to prevent gradient vanishing and/or explosion
in training very deep networks~\cite{Srivastava_2015_Highway}\cite{Srivastava_2015_Training}.

Another family of multi-branch networks follow the bilinear CNN model~\cite{Lin_2015_Bilinear},
which constructs two separate streams to model the co-occurrence of local features.
Formulated as the outer-product of two vectors,
it requires a larger number of parameters and more computational resources than the conventional models to be trained.
An alternative approach is proposed to factorize bilinear models~\cite{Li_2016_Factorized} for visual recognition,
which largely decreases the number of trainable parameters.

All the multi-branch structures are followed by a module to fuse different sources of features.
This can be done by linearly summing them up~\cite{He_2016_Deep}, concatenating them~\cite{Szegedy_2015_Going},
deeply fusing them~\cite{Wang_2016_Deeply},
or using a bilinear~\cite{Lin_2015_Bilinear} or recurrent~\cite{Srivastava_2015_Training} transform.
In this work, we present an extremely simple and efficient approach to enable effective feature ensemble,
which involves introducing a second-order term to apply nonlinear transform in neural responses.
Introducing a second-order operation into neural networks has been studied
in some old-fashioned models~\cite{Goggin_1991_Second}\cite{Kazemy_2007_Second},
but we study this idea in modern deep convolutional networks.

\section{Second-Order Response Transform}
\label{Approach}

\subsection{Formulation}
\label{Approach:Formulation}

Let $\mathbf{x}$ be a set of neural responses at a given layer of a deep neural network.
In practice, $\mathbf{x}$ often appears as a 3D volume.
In a two-branch network structure, $\mathbf{x}$ is fed into two individual modules with different parameters,
and two intermediate data cubes are obtained.
We denote them as $\mathbf{F}_1\!\left(\mathbf{x};\boldsymbol{\theta}_1\right)$
and $\mathbf{F}_2\!\left(\mathbf{x};\boldsymbol{\theta}_2\right)$, respectively.
In the cases without ambiguity,
we write $\mathbf{F}_1\!\left(\mathbf{x}\right)$ and $\mathbf{F}_2\!\left(\mathbf{x}\right)$ in short.
Most often, $\mathbf{F}_1\!\left(\mathbf{x}\right)$ and $\mathbf{F}_2\!\left(\mathbf{x}\right)$ are of the same dimensionality,
and an element-wise operation is used to summarize them into the output set of responses $\mathbf{y}$.

There are some existing examples of two-branch networks,
such as the Maxout network~\cite{Goodfellow_2013_Maxout} and the deep residual network~\cite{He_2016_Deep}.
In Maxout, $\mathbf{F}_1\!\left(\mathbf{x}\right)$ and $\mathbf{F}_2\!\left(\mathbf{x}\right)$
are generated by two individual convolutional layers,
{\em i.e.}, ${\mathbf{F}_m\!\left(\mathbf{x}\right)}={\sigma\!\left[\boldsymbol{\theta}_m^\top\mathbf{x}\right]}$ for ${m}={1,2}$,
where $\boldsymbol{\theta}_m$ is the $m$-th convolutional matrix, $\sigma\!\left[\cdot\right]$ is the activation function,
and an element-wise max operation is performed to fuse them:
${\mathbf{y}^\mathrm{M}}={\max\left\{\mathbf{F}_1\!\left(\mathbf{x}\right),\mathbf{F}_2\!\left(\mathbf{x}\right)\right\}}$.
In a residual module, $\mathbf{F}_1\!\left(\mathbf{x}\right)$ is simply set as an identity mapping ({\em i.e.}, $\mathbf{x}$ itself),
and $\mathbf{F}_2\!\left(\mathbf{x}\right)$ is defined as $\mathbf{x}$ followed by two convolutional operations,
{\em i.e.}, ${\mathbf{F}_2\!\left(\mathbf{x}\right)}=
    {\boldsymbol{\theta}'^\top_2\sigma\!\left[\boldsymbol{\theta}_2^\top\mathbf{x}\right]}$,
and the fusion is performed as linear sum:
${\mathbf{y}^\mathrm{R}}={\mathbf{F}_1\!\left(\mathbf{x}\right)+\mathbf{F}_2\!\left(\mathbf{x}\right)}$.

The core idea of SORT is extremely simple.
We append a second-order term, {\em i.e.} element-wise product, to the linear term, leading to a new fusion strategy:
\begin{equation}
\label{Eqn:SORT}
{\mathbf{y}^\mathrm{S}}={\mathbf{F}_1\!\left(\mathbf{x}\right)+\mathbf{F}_2\!\left(\mathbf{x}\right)+
    g\!\left[\mathbf{F}_1\!\left(\mathbf{x}\right)\odot\mathbf{F}_2\!\left(\mathbf{x}\right)\right]}.
\end{equation}
Here, $\odot$ denotes element-wise product and $g\!\left[\cdot\right]$ is a differentiable function.
The gradient of $\mathbf{y}^\mathrm{S}$ over either $\mathbf{x}$ or $\boldsymbol{\theta}_m$ (${m}={1,2}$) is straightforward.
Note that this modification is very simple yet light-weighted.
Based on a specifically implemented layer in popular deep learning tools such as CAFFE~\cite{Jia_2014_CAFFE},
SORT requires less than $5\%$ additional time in training and testing, meanwhile no extra memory is used.

SORT can be applied to a wide range of network architectures, even if the original structure does not have branches.
In this case, we need to modify each of the original convolutional layers,
{\em i.e.}, ${\mathbf{y}^\mathrm{O}}={\sigma\!\left[\boldsymbol{\theta}^\top\mathbf{x}\right]}$.
We construct two symmetric branches $\mathbf{F}_1\!\left(\mathbf{x}\right)$ and $\mathbf{F}_2\!\left(\mathbf{x}\right)$,
in which the $m$-th branch is defined as ${\mathbf{F}_m\!\left(\mathbf{x}\right)}=
    {\sigma\!\left[\boldsymbol{\theta}'^\top_m\sigma\!\left[\boldsymbol{\theta}_m^\top\mathbf{x}\right]\right]}$.
Then, we perform element-wise fusion~\eqref{Eqn:SORT}
beyond $\mathbf{F}_1\!\left(\mathbf{x}\right)$ and $\mathbf{F}_2\!\left(\mathbf{x}\right)$
by setting $g\!\left[\cdot\right]$ to be an identity mapping function.
Following the idea to reduce the number of parameters~\cite{Simonyan_2015_Very},
we shrink the receptive field size of each convolutional kernel in $\boldsymbol{\theta}_m$
from $k\times k$ to $\left\lfloor\frac{1}{2}\left(k+1\right)\right\rfloor\times\left\lfloor\frac{1}{2}\left(k+1\right)\right\rfloor$.
With two cascaded convolutional layers and $k$ being an odd number,
the overall receptive field size of each neuron in the output layer remains unchanged.
As we shall see in experiments, the branched structure works much better than the original structure,
and SORT consistently boosts the recognition performance beyond the improved baseline.

Another straightforward application of SORT lies in the family of deep residual networks~\cite{He_2016_Deep}.
Note that residual networks are already equipped with two-branch structures,
{\em i.e.}, the input signal $\mathbf{x}$ is followed by an identity mapping and the neural response after two convolutions.
As a direct variant of~\eqref{Eqn:SORT},
SORT modifies the original fusion function from ${\mathbf{y}^\mathrm{R}}={\mathbf{x}+\mathbf{F}\!\left(\mathbf{x}\right)}$ to
${\mathbf{y}^\mathrm{S}}={\mathbf{x}+\mathbf{F}\!\left(\mathbf{x}\right)+
    \sqrt{\mathbf{x}\odot\mathbf{F}\!\left(\mathbf{x}\right)+\varepsilon}}$.
Here ${\varepsilon}={10^{-4}}$ is a small floating point number to avoid numerical instability in gradient computation.
Note that in the residual networks,
elements in either $\mathbf{x}$ or $\mathbf{F}\!\left(\mathbf{x}\right)$ may be negative~\cite{He_2016_Identity},
and we perform a ReLU activation on it before computing the product term.
Thus, the exact form of SORT in this case is
${\mathbf{y}^\mathrm{S}}={\mathbf{x}+\mathbf{F}\!\left(\mathbf{x}\right)+
    \sqrt{\sigma\!\left[\mathbf{x}\right]\odot\sigma\!\left[\mathbf{F}\!\left(\mathbf{x}\right)\right]+\varepsilon}}$.
Similarly, SORT does not change the receptive field size of an output neuron.

\subsection{Cross-Branch Response Propagation}
\label{Approach:Propagation}

We first discuss the second-order term.
According to our implementation, all the numbers fed into element-wise product are non-negative,
{\em i.e.}, $\forall i$, ${F_{1,i}\!\left(\mathbf{x}\right)}\geqslant{0}$ and ${F_{2,i}\!\left(\mathbf{x}\right)}\geqslant{0}$.
Therefore, the second-order term is either $0$ or a positive value
(when both $F_{1,i}\!\left(\mathbf{x}\right)$ and $F_{2,i}\!\left(\mathbf{x}\right)$ are positive).
Consider two input pairs,
{\em i.e.}, ${\left(F_{1,i}\!\left(\mathbf{x}\right),F_{2,i}\!\left(\mathbf{x}\right)\right)}={\left(a,0\right)}$
or ${\left(F_{1,i}\!\left(\mathbf{x}\right),F_{2,i}\!\left(\mathbf{x}\right)\right)}={\left(a_1,a_2\right)}$ where ${a_1+a_2}={a}$.
In the former case we have ${y_i^\mathrm{S}}={a}$, but in the latter case we have ${y_i^\mathrm{S}}={a+a_1\times a_2}$.
The extra term, {\em i.e.}, $a_1\times a_2$, is large when $a_1$ and $a_2$ are close, {\em i.e.}, $\left|a_1-a_2\right|$ is small.
We explain it as facilitating the {\em consistent} responses,
{\em i.e.}, we reward the indices on which two branches have similar response values.

We also note that SORT leads to an improved way of gradient back-propagation.
Since there exists a dyadic term $\mathbf{F}_1\!\left(\mathbf{x};\boldsymbol{\theta}_1\right)\odot
    \mathbf{F}_2\!\left(\mathbf{x};\boldsymbol{\theta}_2\right)$,
the gradient of $\mathbf{y}^\mathrm{S}$ with respect to either one
in $\boldsymbol{\theta}_1$ and $\boldsymbol{\theta}_2$ is related to another.
Thus, when the parameter $\boldsymbol{\theta}_1$ needs to be updated,
the gradient $\frac{\partial L}{\partial\boldsymbol{\theta}_1}$ is directly related to $\mathbf{F}_2\!\left(\mathbf{x}\right)$:
\begin{equation}
\label{Eqn:Gradient}
{\frac{\partial L}{\partial\boldsymbol{\theta}_1}}=
    {\left(\frac{\partial L}{\partial\mathbf{y}^\mathrm{S}}\right)^\top\cdot
    \left[1+\mathbf{F}_2\!\left(\mathbf{x};\boldsymbol{\theta}_2\right)\right]^\top\cdot
    {\frac{\partial\mathbf{F}_1\!\left(\mathbf{x};\boldsymbol{\theta}_1\right)}{\partial\boldsymbol{\theta}_1}}},
\end{equation}
and similarly, $\frac{\partial L}{\partial\boldsymbol{\theta}_2}$ is directly related to $\mathbf{F}_1\!\left(\mathbf{x}\right)$.
This prevents the gradients from being shattered as the network goes deep~\cite{Balduzzi_2017_Shattered},
and reduces the risk of structural over-fitting ({\em i.e.}, caused by the increasing number of network layers).
As an example, we train deep residual networks~\cite{He_2016_Deep} with different numbers of layers
on the {\bf SVHN} dataset~\cite{Netzer_2011_Reading}, a relatively simple dataset for street house number recognition.
Detailed experimental settings are illustrated in Section~\ref{Experiments:SmallScale}.
The baseline recognition errors are $2.30\%$ and $2.49\%$ for the $20$-layer and $56$-layer networks, respectively,
while these numbers become $2.26\%$ and $2.19\%$ after SORT is applied.
SORT consistently improves the recognition rate,
and the gain becomes more significant when a deeper network architecture is used.

In summary, SORT allows the network to consider cross-branch information in both forward-propagation and back-propagation.
This strategy improves the reliability of neural responses, as well as the numerical stability in gradient computation.

\subsection{Global Network Nonlinearity}
\label{Approach:Nonlinearity}

\renewcommand{\figurewidth}{8.0cm}
\begin{figure}
\begin{center}
    \includegraphics[width=\figurewidth]{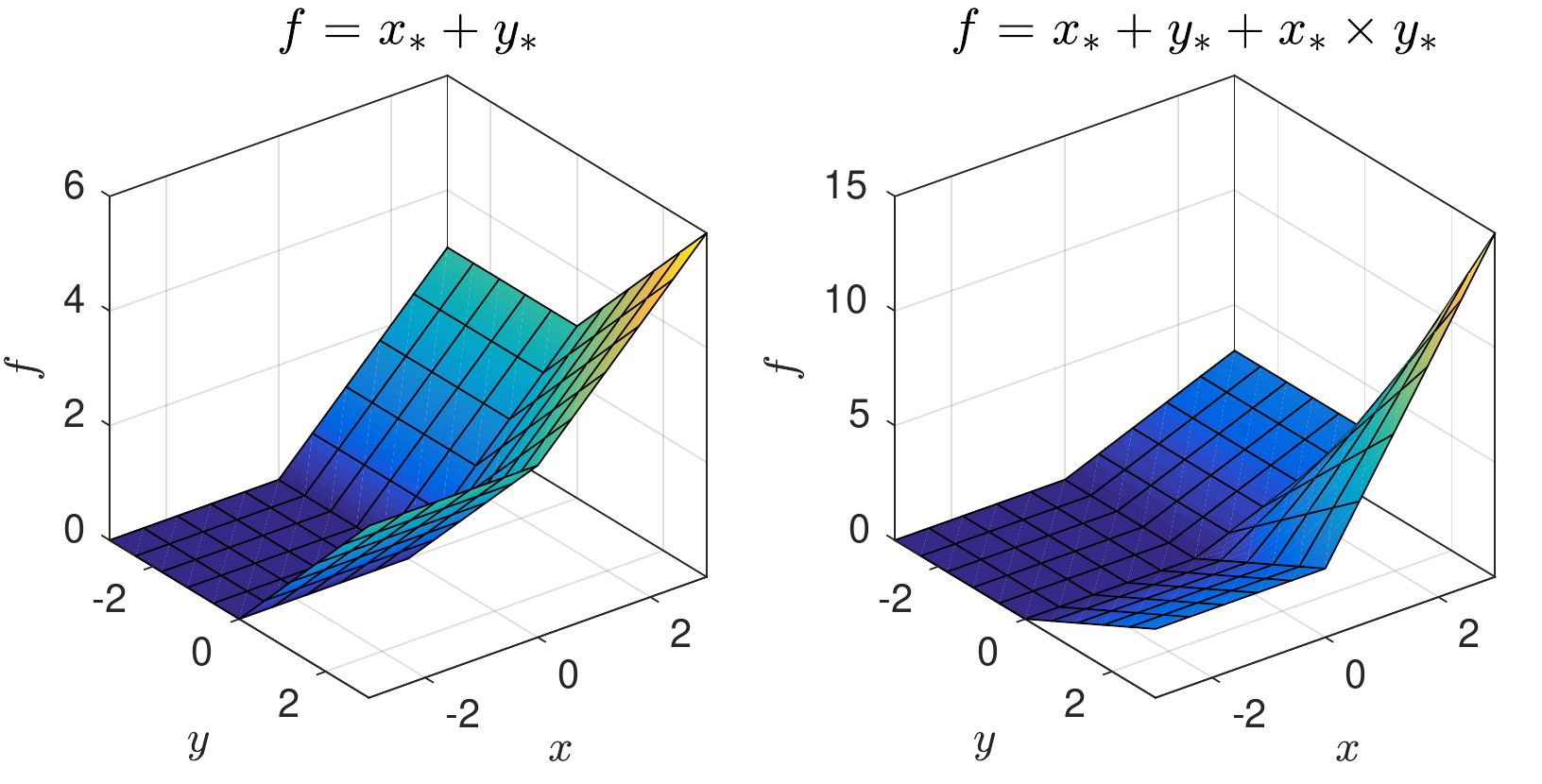}
\end{center}
\caption{
    Comparison of different response transform functions.
    The second-order operation produces nonlinearity in a 2D subset.
    Here, ${x_\ast}\doteq{\max\left\{x,0\right\}}$ and ${y_\ast}\doteq{\max\left\{y,0\right\}}$.
}
\label{Fig:Nonlinearity}
\end{figure}

Nonlinearity makes the major contribution to the representation ability of deep neural networks~\cite{Jarrett_2009_What}.
State-of-the-art networks are often equipped with sigmoid or ReLU activation~\cite{Nair_2010_Rectified} and/or max-pooling layers,
and we argue that the proposed second-order term is a better choice.
To this end, we consider two functions ${f_1\!\left(x,y\right)}={x_\ast+y_\ast}$ and
${f_2\!\left(x,y\right)}={x_\ast+y_\ast+x_\ast\times y_\ast}$,
where ${x_\ast}\doteq{\max\left\{x,0\right\}}$ and ${y_\ast}\doteq{\max\left\{y,0\right\}}$ are responses after ReLU activation.
If the second-order term is not involved, we obtain a piecewise linear function $f_1\!\left(x,y\right)$,
which means that nonlinearity only appears in several 1D subspaces of the 2D plane $\mathbb{R}^2$.
By adding the second-order term, nonlinearity exists in ${\mathbb{R}_\ast^2}\doteq{\left[0,+\infty\right)^2}$
(see Figure~\ref{Fig:Nonlinearity}).

\renewcommand{\colwidthA}{0.7cm}
\renewcommand{\colwidthB}{1.2cm}
\begin{table}
\centering
\begin{tabular}{|C{\colwidthA}|C{\colwidthA}|C{\colwidthA}||R{\colwidthB}|R{\colwidthB}|R{\colwidthB}|}
\hline
$+$        & $\max$     & $\odot$    & {\bf LeNet}      & {\bf BigNet}     & {\bf ResNet}     \\
\hline\hline
\checkmark &            &            & $11.10$          & $ 6.86$          & $ 7.60$          \\
\hline
           & \checkmark &            & $11.07$          & $ 7.01$          & $ 7.55$          \\
\hline
           &            & \checkmark & $11.03$          & $    -$          & $    -$          \\
\hline
\checkmark & \checkmark &            & $11.02$          & $ 6.90$          & $ 7.63$          \\
\hline
\checkmark &            & \checkmark & $\mathbf{10.34}$ & $ 6.60$          & $\mathbf{ 7.14}$ \\
\hline
           & \checkmark & \checkmark & $10.39$          & $\mathbf{ 6.57}$ & $ 7.44$          \\
\hline
\checkmark & \checkmark & \checkmark & $10.80$          & $ 6.65$          & $ 7.90$          \\
\hline
\end{tabular}
\caption{
    Recognition error rate ($\%$) on the {\bf CIFAR10} dataset with different fusion strategies.
    Here, $+$, $\max$ and $\odot$ denote three dyadic operators,
    and multiple checkmarks in one row means to sum up the results produced by the corresponding operators.
    Sometimes, using the second-order terms alone results in non-convergence (denoted by $-$).
    All these numbers are averaged over $3$ individual runs, with standard deviations of $0.04\%$--$0.08\%$.
}
\label{Tab:Comparison}
\end{table}

Summarizing the cues above (cross-branch propagation and nonlinearity)
leads to adding a second-order term which involves neural responses from both branches.
Hence, $\mathbf{F}_1\odot\mathbf{F}_2$ is a straightforward and simple choice.
We point out that an alternative choice of second-term nonlinearity is the square term,
{\em i.e.}, $\mathbf{F}_1^2\left(\mathbf{x}\right)$, where $\cdot^2$ denotes the element-wise operation.
but we do not suggest this option, since this does not allow cross-branch response propagation.
As a side note, an element-wise product term behaves similarly to a logical-and term,
which is verified effective in learning feature representations in neural networks~\cite{Mansour_1992_Learning}.

We experimentally verify the effectiveness of nonlinearity by considering three fusion strategies,
{\em i.e.}, $\mathbf{F}_1\!\left(\mathbf{x}\right)+\mathbf{F}_2\!\left(\mathbf{x}\right)$,
$\max\left\{\mathbf{F}_1\!\left(\mathbf{x}\right),\mathbf{F}_2\!\left(\mathbf{x}\right)\right\}$
and $\sqrt{\mathbf{F}_1\!\left(\mathbf{x}\right)\odot\mathbf{F}_2\!\left(\mathbf{x}\right)}$.
To compare their performance, we apply different fusion strategies on different networks,
and evaluate them on the {\bf CIFAR10} dataset (detailed settings are elaborated in Section~\ref{Experiments:SmallScale}).
Various combinations lead to different recognition results, which are summarized in Table~\ref{Tab:Comparison}.

We first note that the second-order operator $\odot$ shall not be used alone,
since this often leads to non-convergence especially in those very deep networks,
{\em e.g.}, {\bf BigNet} ($19$ layers) and {\bf ResNet} ($20$ layers).
The learning curves in Figure~\ref{Fig:SmallDatasetCurves} also provide evidences to this point.
It is well acknowledged that first-order terms are able to provide numerical stability,
and help the training process converge~\cite{Nair_2010_Rectified} compared to some saturable activation functions such as sigmoid.
On the other hand, when the second-order term is appended to either $+$ or $\max$, the recognition error is significantly decreased,
which suggests that adding higher-order terms indeed increases the network representation ability,
which helps to better depict the complicated feature space and achieve higher recognition rates.
Missing either the first-order or second-order term harms the recognition accuracy of the deep network,
thus we suggest to use a combination of linear and nonlinear terms in all the later experiments.
In practice, we choose the linear sum mainly because it allows both branches to get trained in back-propagation,
while the max operator only updates half of the parameters at each time.
In addition, the max operator does not reward consistent responses as the second-order term does.

\subsection{Relationship to Other Work}
\label{Approach:Relationship}

We note that some previous work also proposed to use a second-order term in network training.
For example, the bilinear CNN~\cite{Lin_2015_Bilinear} computes the outer-product of neural responses from two individual networks
to capture feature co-occurrence at the same spatial positions.
However, this operation often requires heavy time and memory overheads,
as it largely increases the dimensionality of the feature vector, and consequently the number of trainable parameters.
Training a bilinear CNN is often slow, even in the improved versions~\cite{Gao_2016_Compact}\cite{Li_2016_Factorized}.
In comparison, the extra computation brought by SORT is merely ignorable ($<5\%$).
We evaluate~\cite{Lin_2015_Bilinear} and~\cite{Gao_2016_Compact} on the {\bf CIFAR10} dataset.
Using {\bf BigNet*}~\cite{Nagadomi_2014_Kaggle} as the backbone (see Section~\ref{Experiments:SmallScale:Settings}),
the error rates of~\cite{Lin_2015_Bilinear}, \cite{Gao_2016_Compact} and SORT are $7.17\%$, $8.01\%$ and $6.81\%$,
and every $20$ iterations take $3.7\mathrm{s}$, $16.5\mathrm{s}$ and $2.1\mathrm{s}$, respectively.
Compared with the baseline, bilinear pooling requires heavier computation and reports even worse results.
This was noted in the original paper~\cite{Lin_2015_Bilinear},
which shows that good initialization and careful fine-tuning are required,
and therefore it was not designed for training-from-scratch.

In a spatial transformer network~\cite{Jaderberg_2015_Spatial},
the product operator is used to apply an affine transform on the neural responses.
In some attention-based models~\cite{Chen_2016_Attention},
product operations are also used to adjust the intensity of neurons according to the spatial weights.
We point out that SORT is generalized.
Due to its simplicity and efficiency, it can be applied to many different network structures.

SORT is also related to the gating function used in recurrent neural network cells
such as the long short-term memory (LSTM)~\cite{Hochreiter_1997_Long} or the gated recurrent unit (GRU)~\cite{Chung_2014_Empirical}.
There, element-wise product is used at each time step to regularize the memory cell and the hidden state.
This operation has also been explored in computer vision~\cite{Srivastava_2015_Highway} to facilitate very deep network training.
In comparison, our method introduces second-order transform without adding new parameters,
whereas the second-order terms in~\cite{Hochreiter_1997_Long} or~\cite{Srivastava_2015_Highway}
require extra parameters for every newly-added gate.

\section{Experiments}
\label{Experiments}

We apply the second-order response transform (SORT) to several popular network architectures,
including chain-styled networks ({\bf LeNet}, {\bf BigNet} and {\bf AlexNet}) and two variants of deep residual networks.
We verify significant accuracy gain over a wide range of visual recognition tasks.

\subsection{Small-Scale Experiments}
\label{Experiments:SmallScale}

\subsubsection{Settings}
\label{Experiments:SmallScale:Settings}

Three small-scale datasets are used in this section.
Among them, the {\bf CIFAR10} and {\bf CIFAR100} datasets~\cite{Krizhevsky_2009_Learning}
are subsets drawn from the $80$-million tiny image database~\cite{Torralba_2008_80}.
Each set contains $50\rm{,}000$ training samples and $10\rm{,}000$ testing samples, and each sample is a $32\times32$ RGB image.
In both datasets, training and testing samples are uniformly distributed over all the categories
({\bf CIFAR10} contains $10$ basic classes, and {\bf CIFAR100} has $100$ where the visual concepts are defined at a finer level).
The {\bf SVHN} dataset~\cite{Netzer_2011_Reading} is a larger collection for digit recognition,
{\em i.e.}, there are $73\rm{,}257$ training samples, $26\rm{,}032$ testing samples, and $531\rm{,}131$ extra training samples.
Each sample is also a $32\times32$ RGB image.
We preprocess the data as in the previous literature~\cite{Netzer_2011_Reading},
{\em i.e.}, selecting $400$ samples per category from the training set as well as $200$ samples per category from the extra set,
using these $6\rm{,}000$ images for validation, and the remaining $598\rm{,}388$ images as training samples.
We also use local contrast normalization (LCN) for data preprocessing~\cite{Goodfellow_2013_Maxout}.

Four baseline network architectures are evaluated.

\begin{itemize}
\item {\bf LeNet}~\cite{LeCun_1998_Gradient} is a relatively shallow network with $3$ convolutional layers,
$3$ pooling layers and $2$ fully-connected layers.
All the convolutional layers have $5\times5$ kernels,
and the input cube is zero-padded by a width of $2$ so that the spatial resolution of the output remains unchanged.
After each convolution including the first fully-connected layer,
a nonlinear function known as ReLU~\cite{Nair_2010_Rectified} is used for activating the neural responses.
This common protocol will be used in all the network structures.
The pooling layers have $3\times3$ kernels, and a spatial stride of $2$.
We apply three training sections with learning rates of $10^{-2}$, $10^{-3}$ and $10^{-4}$,
and $60\mathrm{K}$, $5\mathrm{K}$, and $5\mathrm{K}$ iterations, respectively.
\item A so-called {\bf BigNet} is trained as a deeper chain-styled network.
There are $10$ convolutional layers, $3$ pooling layers and $3$ fully-connected layers in this architecture.
The design of {\bf BigNet} is similar to {\bf VGGNet}~\cite{Simonyan_2015_Very},
in which small convolutional kernels ($3\times3$) are used and the depth is increased.
Following~\cite{Nagadomi_2014_Kaggle},
we apply four training sections with learning rates of $10^{-1}$, $10^{-2}$, $10^{-3}$ and $10^{-4}$,
and $60\mathrm{K}$, $30\mathrm{K}$, $20\mathrm{K}$ and $10\mathrm{K}$ iterations, respectively.
\item The deep residual network ({\bf ResNet})~\cite{He_2016_Deep} brings significant performance boost beyond chain-styled networks.
We follow the original work~\cite{He_2016_Deep} to define network architectures with different numbers of layers,
which are denoted as {\bf ResNet-20}, {\bf ResNet-32} and {\bf ResNet-56}, respectively.
These architectures differ from each other in the number of residual blocks used in each stage.
Batch normalization is applied after each convolution to avoid numerical instability in this very deep network.
Following the implementation of~\cite{Xu_2016_Residual},
we apply three training sections with learning rates of $10^{-1}$, $10^{-2}$, and $10^{-3}$,
and $32\mathrm{K}$, $16\mathrm{K}$ and $16\mathrm{K}$ iterations, respectively.
\item The wide residual network ({\bf WRN})~\cite{Zagoruyko_2016_Wide} takes the idea
to increase the number of kernels in each layer and decrease the network depth at the same time.
We apply the $28$-layer architecture, denoted as {\bf WRN-28}, which is verified effective in~\cite{Zagoruyko_2016_Wide}.
Following the same implementation of the original {\bf ResNet}s,
we apply three training sections with learning rates of $10^{-1}$, $10^{-2}$ and $10^{-3}$,
and $32\mathrm{K}$, $16\mathrm{K}$, and $16\mathrm{K}$ iterations, respectively.
\end{itemize}

In all the networks, the mini-batch size is fixed as $100$.
Note that both {\bf LeNet} and {\bf BigNet} are chain-styled networks.
Using the details illustrated in Section~\ref{Approach:Formulation},
we replace each convolutional layer using a two-branch, two-layer module with smaller kernels.
This leads to deeper and more powerful networks, and we append an asterisk ({\bf *}) after the original networks to denote them.
SORT is applied to the modified network structure by appending element-wise product to linear sum.

\subsubsection{Results}
\label{Experiments:SmallScale:Results}

\renewcommand{\colwidth}{1.0cm}
\begin{table}
\centering
\begin{tabular}{|l||R{\colwidth}|R{\colwidth}|R{\colwidth}|}
\hline
Network                                          & {\bf CF10}       & {\bf CF100}      & {\bf SVHN}       \\
\hline\hline
Lee {\em et.al}~\cite{Lee_2015_Deeply}           & $ 7.97$          & $34.57$          & $ 1.92$          \\
\hline
Liang {\em et.al}~\cite{Liang_2015_Recurrent}    & $ 7.09$          & $31.75$          & $ 1.77$          \\
\hline
Lee {\em et.al}~\cite{Lee_2016_Generalizing}     & $ 6.05$          & $32.37$          & $ 1.69$          \\
\hline
Wang {\em et.al}~\cite{Wang_2016_Deeply}         & $ 5.87$          & $27.01$          & $    -$          \\
\hline
Zagoruyko {\em et.al}~\cite{Zagoruyko_2016_Wide} & $ 5.37$          & $24.53$          & $ 1.85$          \\
\hline
Xie {\em et.al}~\cite{Xie_2016_Geometric}        & $ 5.31$          & $25.01$          & $ 1.67$          \\
\hline
Huang {\em et.al}~\cite{Huang_2016_Deep}         & $ 5.25$          & $24.98$          & $ 1.75$          \\
\hline
Huang {\em et.al}~\cite{Huang_2017_Densely}      & $\mathbf{ 3.74}$ & $\mathbf{19.25}$ & $\mathbf{ 1.59}$ \\
\hline\hline
{\bf LeNet}                                      & $14.37$          & $43.83$          & $ 4.00$          \\
\hline
{\bf LeNet*}                                     & $11.16$          & $36.84$          & $ 2.65$          \\
\hline
{\bf LeNet*}-SORT                                & $\mathbf{10.41}$ & $\mathbf{34.67}$ & $\mathbf{ 2.47}$ \\
\hline\hline
{\bf BigNet}                                     & $ 7.55$          & $30.47$          & $ 2.21$          \\
\hline
{\bf BigNet*}                                    & $ 6.92$          & $29.43$          & $ 2.17$          \\
\hline
{\bf BigNet*}-SORT                               & $\mathbf{ 6.81}$ & $\mathbf{28.10}$ & $\mathbf{ 2.12}$ \\
\hline\hline
{\bf ResNet-20}                                  & $ 7.72$          & $31.80$          & $ 2.30$          \\
\hline
{\bf ResNet-20}-SORT                             & $\mathbf{ 7.35}$ & $\mathbf{31.65}$ & $\mathbf{ 2.26}$ \\
\hline\hline
{\bf ResNet-32}                                  & $ 6.83$          & $30.28$          & $ 2.54$          \\
\hline
{\bf ResNet-32}-SORT                             & $\mathbf{ 6.33}$ & $\mathbf{29.61}$ & $\mathbf{ 2.22}$ \\
\hline\hline
{\bf ResNet-56}                                  & $ 6.30$          & $28.25$          & $ 2.49$          \\
\hline
{\bf ResNet-56}-SORT                             & $\mathbf{ 5.50}$ & $\mathbf{26.76}$ & $\mathbf{ 2.19}$ \\
\hline\hline
{\bf WRN-28}                                     & $ 4.81$          & $21.90$          & $ 1.93$          \\
\hline
{\bf WRN-28}-SORT                                & $\mathbf{ 4.48}$ & $\mathbf{21.52}$ & $\mathbf{ 1.48}$ \\
\hline
\end{tabular}
\caption{
    Recognition error rate ($\%$) on small datasets and different network architectures.
    All the numbers are averaged over $3$ individual runs, and the standard deviation is often less than $0.08\%$.
}
\label{Tab:SmallDatasets}
\end{table}

Results are summarized in Table~\ref{Tab:SmallDatasets}.
One can observe that SORT boosts the performance of all network architectures consistently.
On both {\bf LeNet} and {\bf BigNet}, we observe significant accuracy gain brought by
replacing of each convolutional layer as a two-branch module.
SORT further improves recognition accuracy by using a more effective fusion function.
In addition, we observe more significant accuracy gain when the network goes deeper.
For example, on the $20$-layer {\bf ResNet},
the relative error rate drops are $4.79$, $0.47\%$ and $1.74\%$ for {\bf CIFAR10}, {\bf CIFAR100}) and {\bf SVHN},
and these numbers become much bigger ($12.70$, $5.27\%$ and $12.05\%$, respectively) on the $56$-layer {\bf ResNet}.
This verifies our hypothesis in Section~\ref{Approach:Propagation},
that SORT alleviates the shattered gradient problem and helps training very deep networks more efficiently.
Especially, based on {\bf WRN-28}, one of the state-of-the-art structures,
SORT reduces the recognition error rate of {\bf SVHN} from $1.93\%$ to $1.48\%$,
giving a relatively $23.32\%$ error drop, meanwhile achieving the new state-of-the-art
(the previous record is $1.59\%$~\cite{Huang_2017_Densely}).
All these results suggest the usefulness of the second-order term in visual recognition.

\subsubsection{Discussions}
\label{Experiments:SmallScale:Discussions}

\renewcommand{\scatterwidth}{5.6cm}
\renewcommand{\subfigurehkern}{0.1cm}
\renewcommand{\subfigurevkern}{0.15cm}
\begin{figure*}
\begin{center}
\begin{minipage}{\scatterwidth}
\centering
    \includegraphics[width=\scatterwidth]{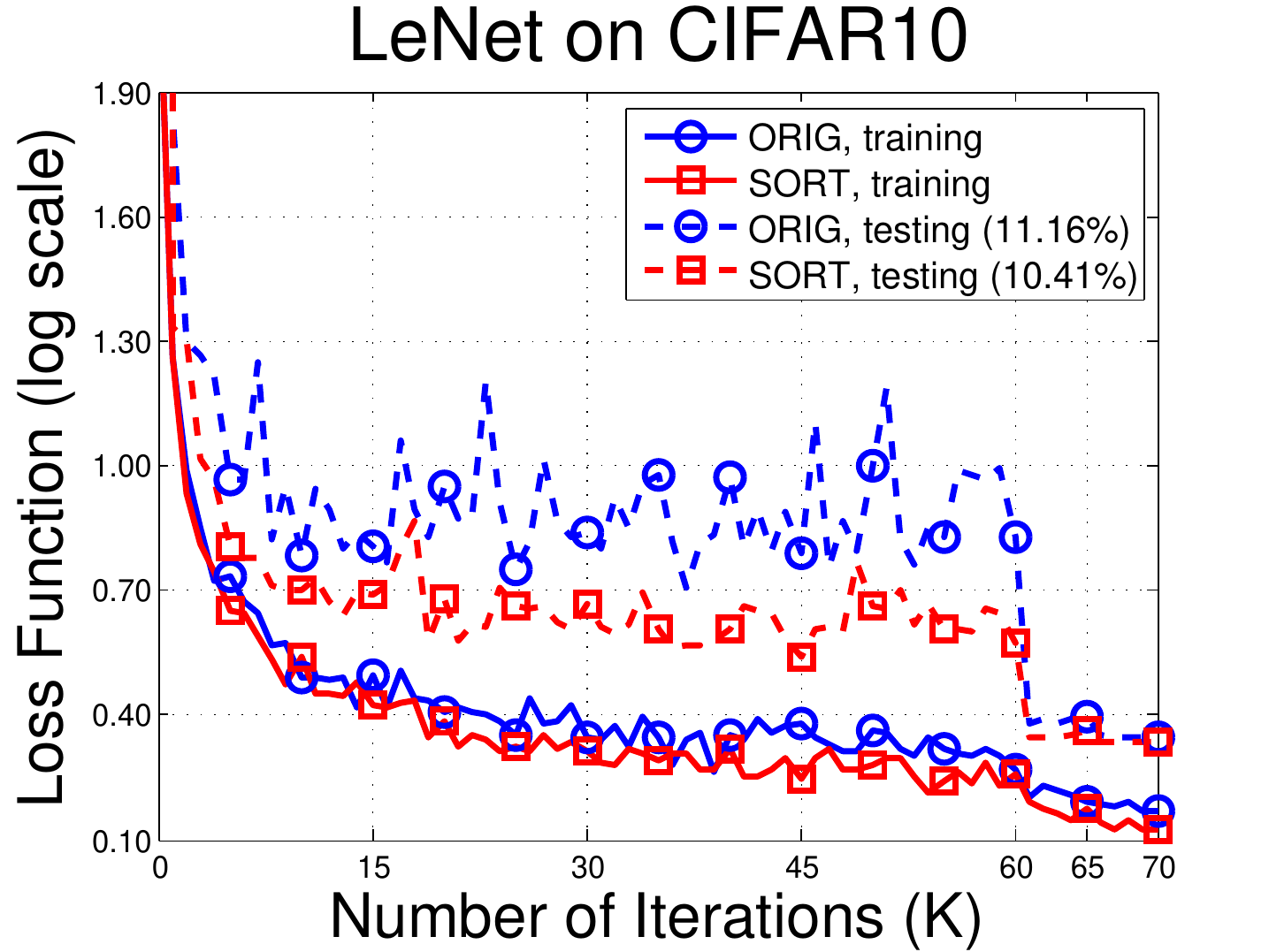}
\end{minipage}
\hspace{\subfigurehkern}
\begin{minipage}{\scatterwidth}
\centering
    \includegraphics[width=\scatterwidth]{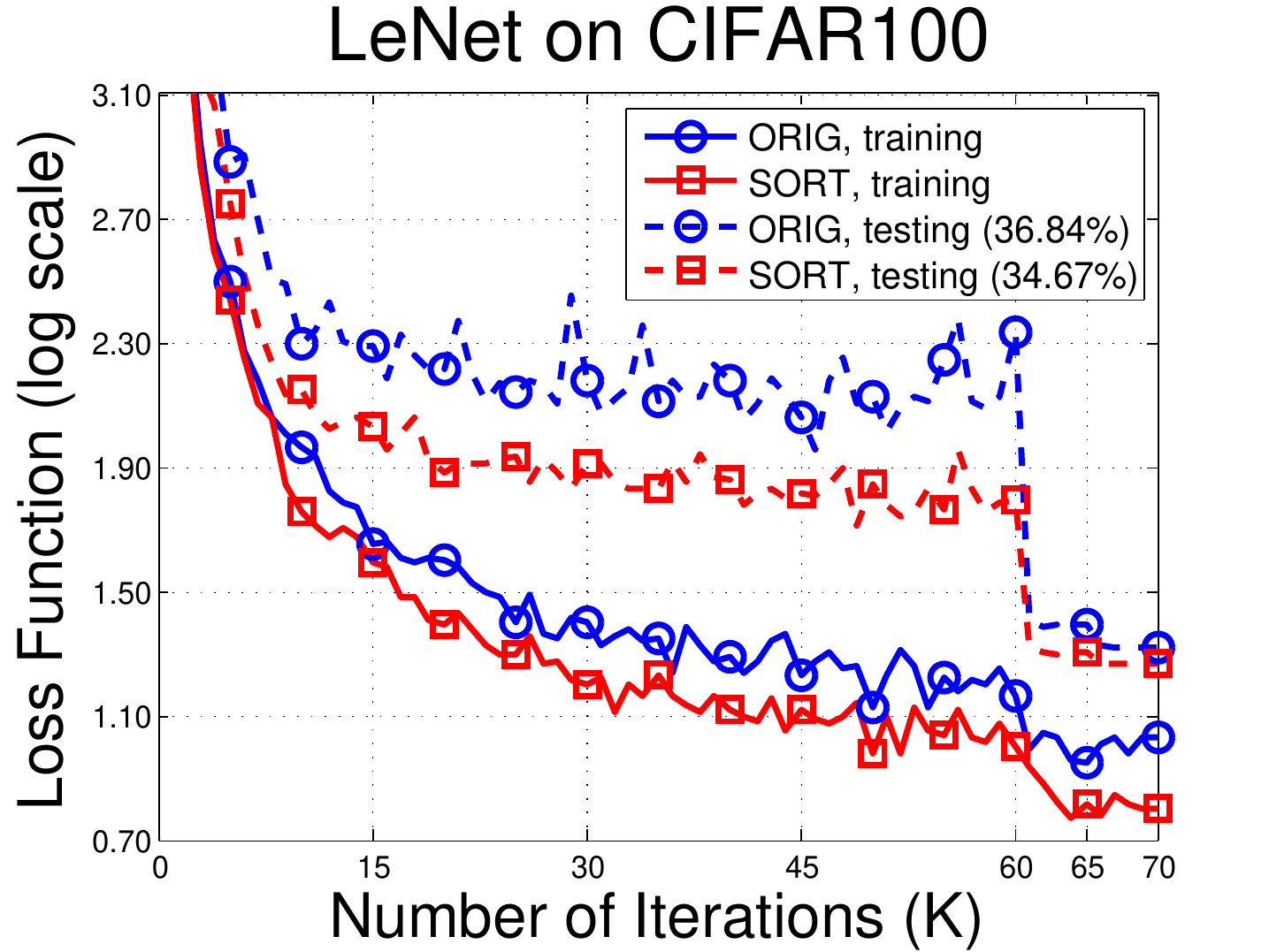}
\end{minipage}
\hspace{\subfigurehkern}
\begin{minipage}{\scatterwidth}
\centering
    \includegraphics[width=\scatterwidth]{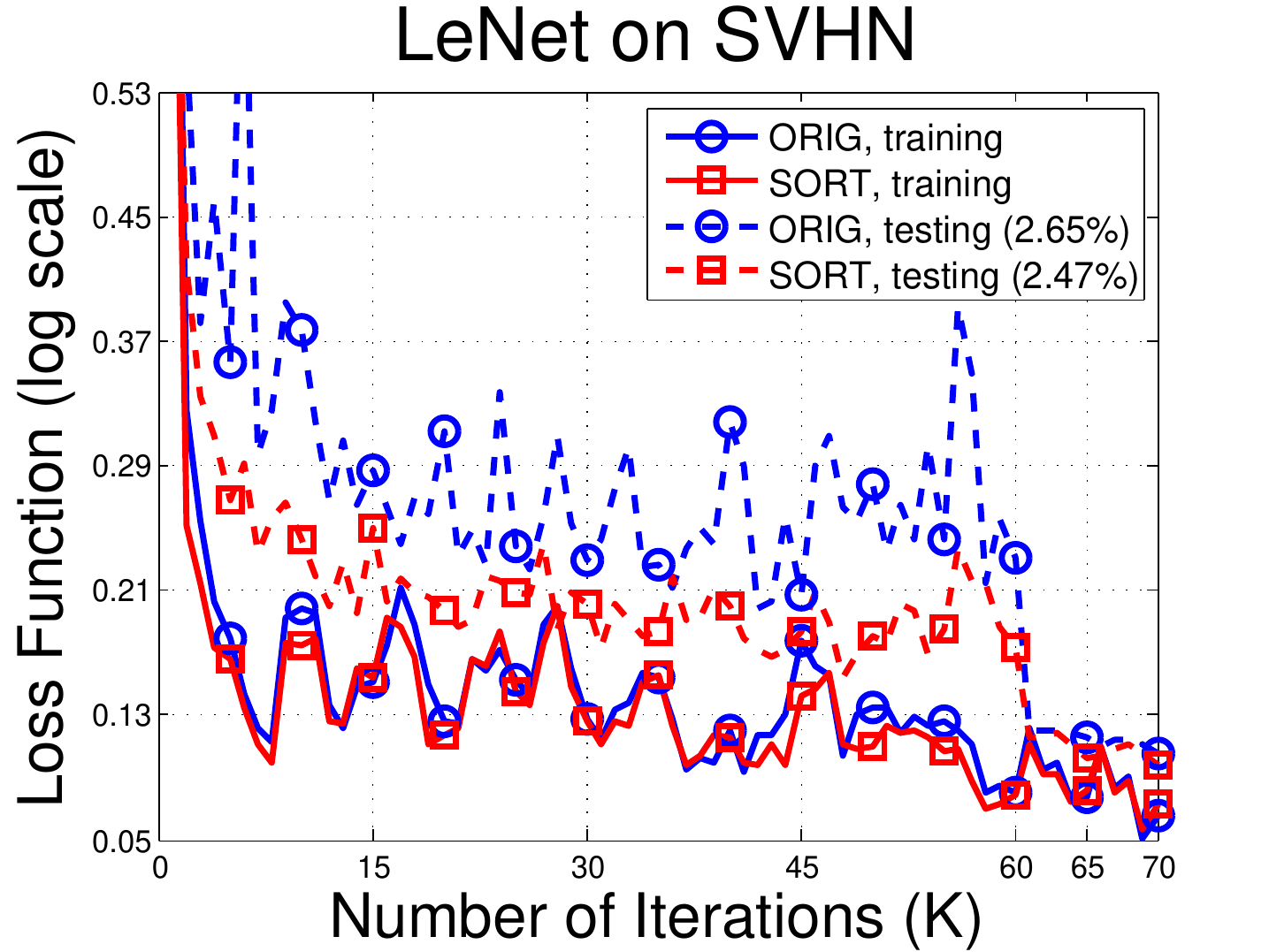}
\end{minipage} \\
\vspace{\subfigurevkern}
\begin{minipage}{\scatterwidth}
\centering
    \includegraphics[width=\scatterwidth]{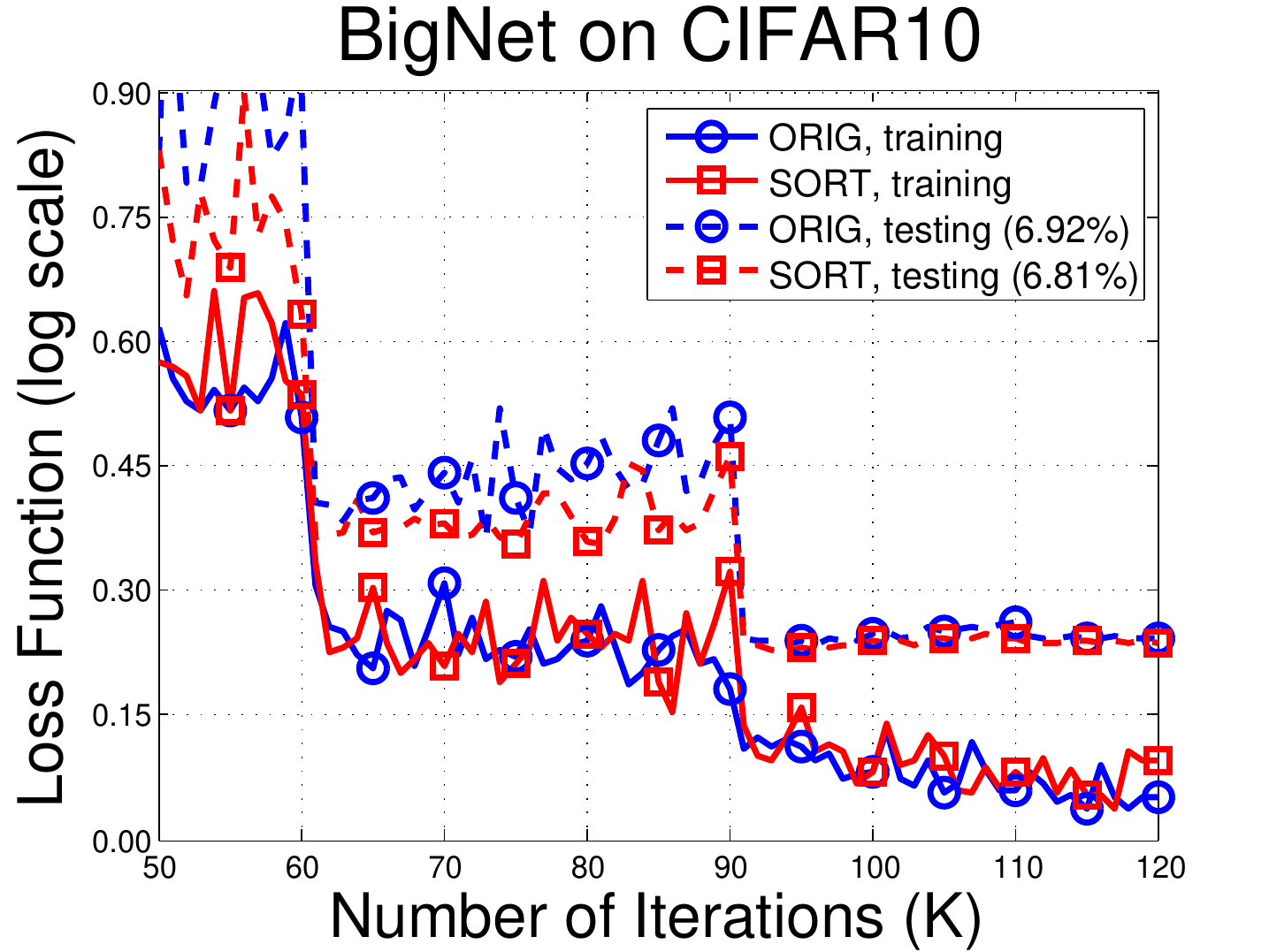}
\end{minipage}
\hspace{\subfigurehkern}
\begin{minipage}{\scatterwidth}
\centering
    \includegraphics[width=\scatterwidth]{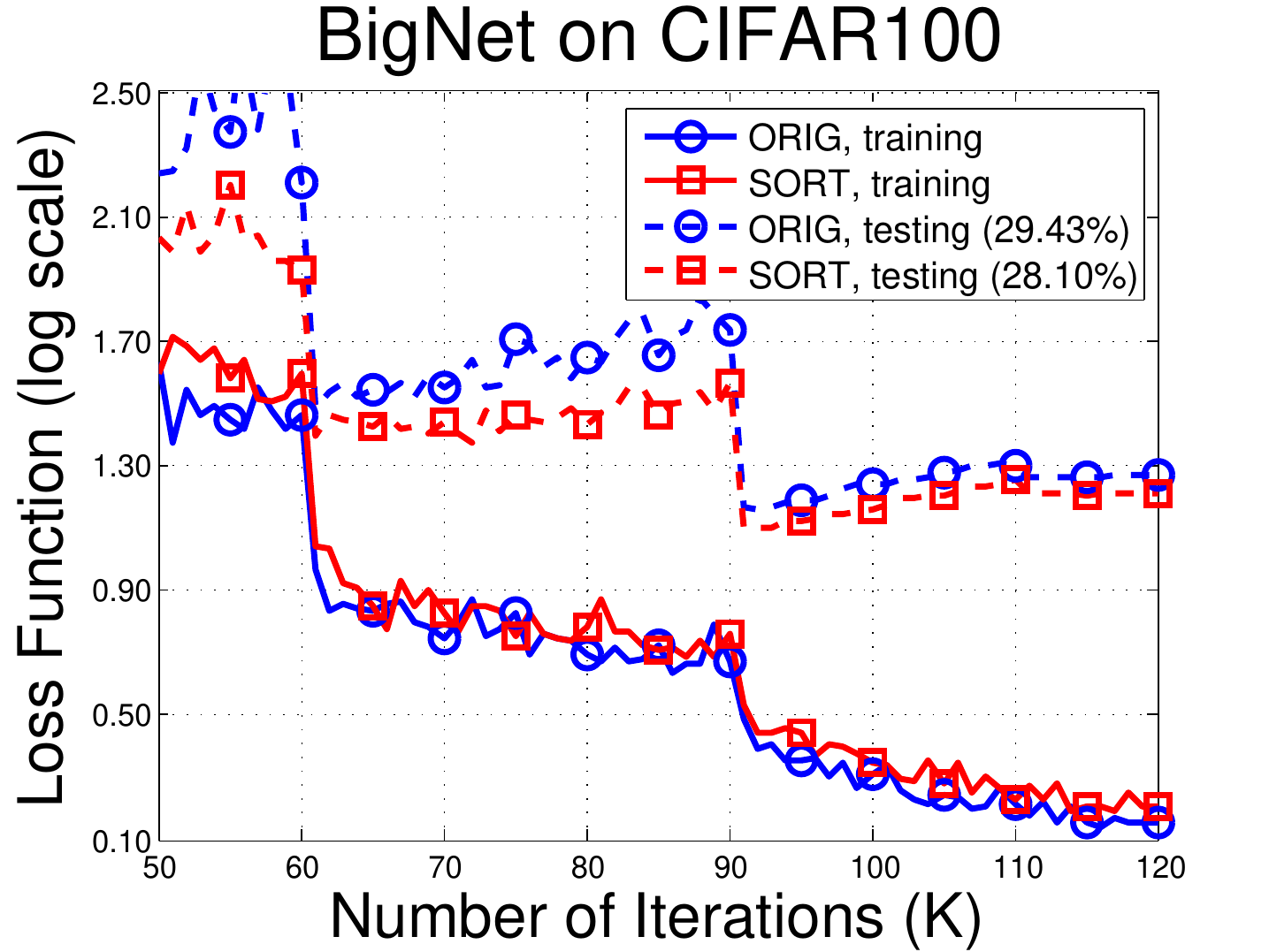}
\end{minipage}
\hspace{\subfigurehkern}
\begin{minipage}{\scatterwidth}
\centering
    \includegraphics[width=\scatterwidth]{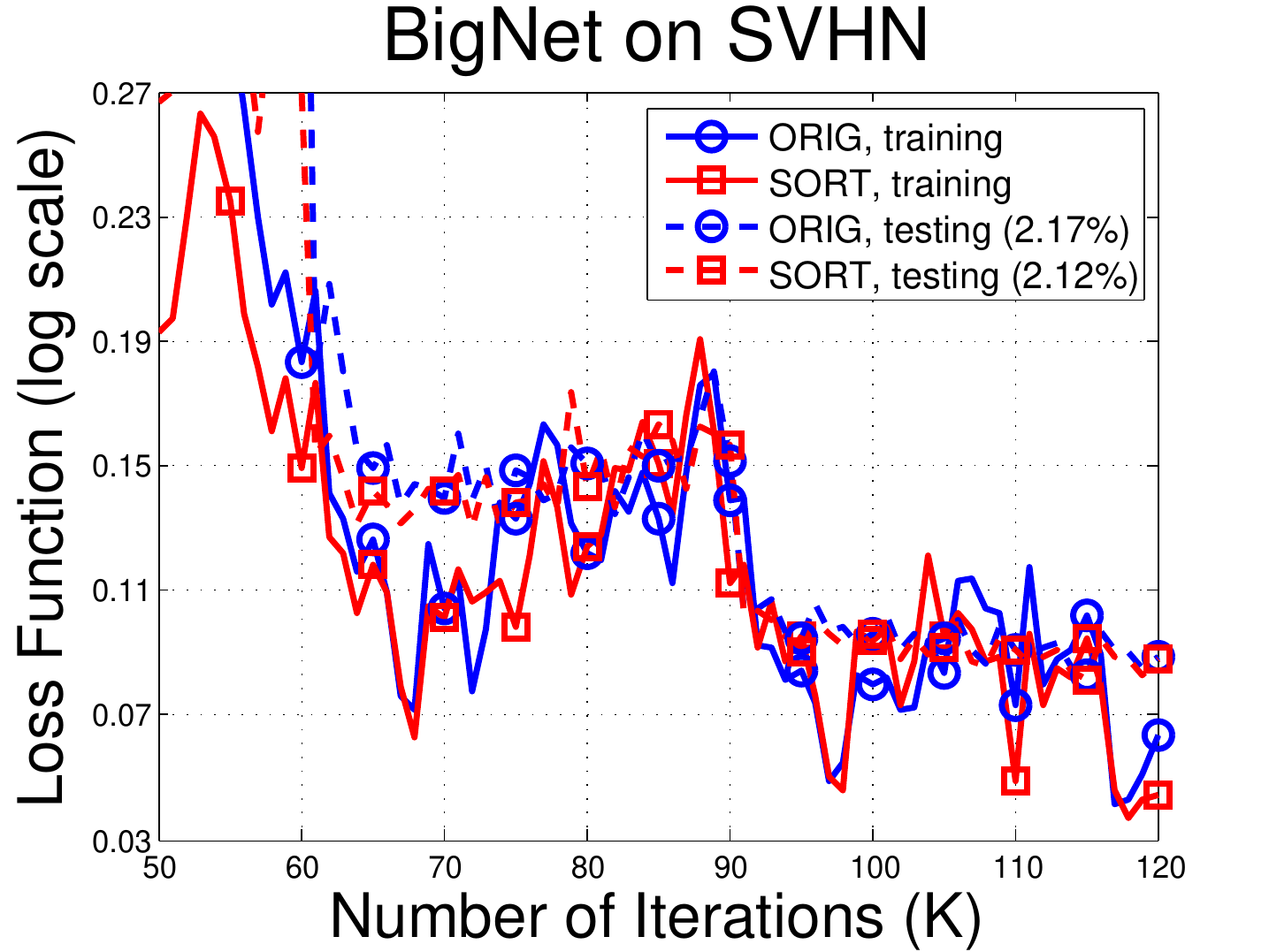}
\end{minipage} \\
\vspace{\subfigurevkern}
\begin{minipage}{\scatterwidth}
\centering
    \includegraphics[width=\scatterwidth]{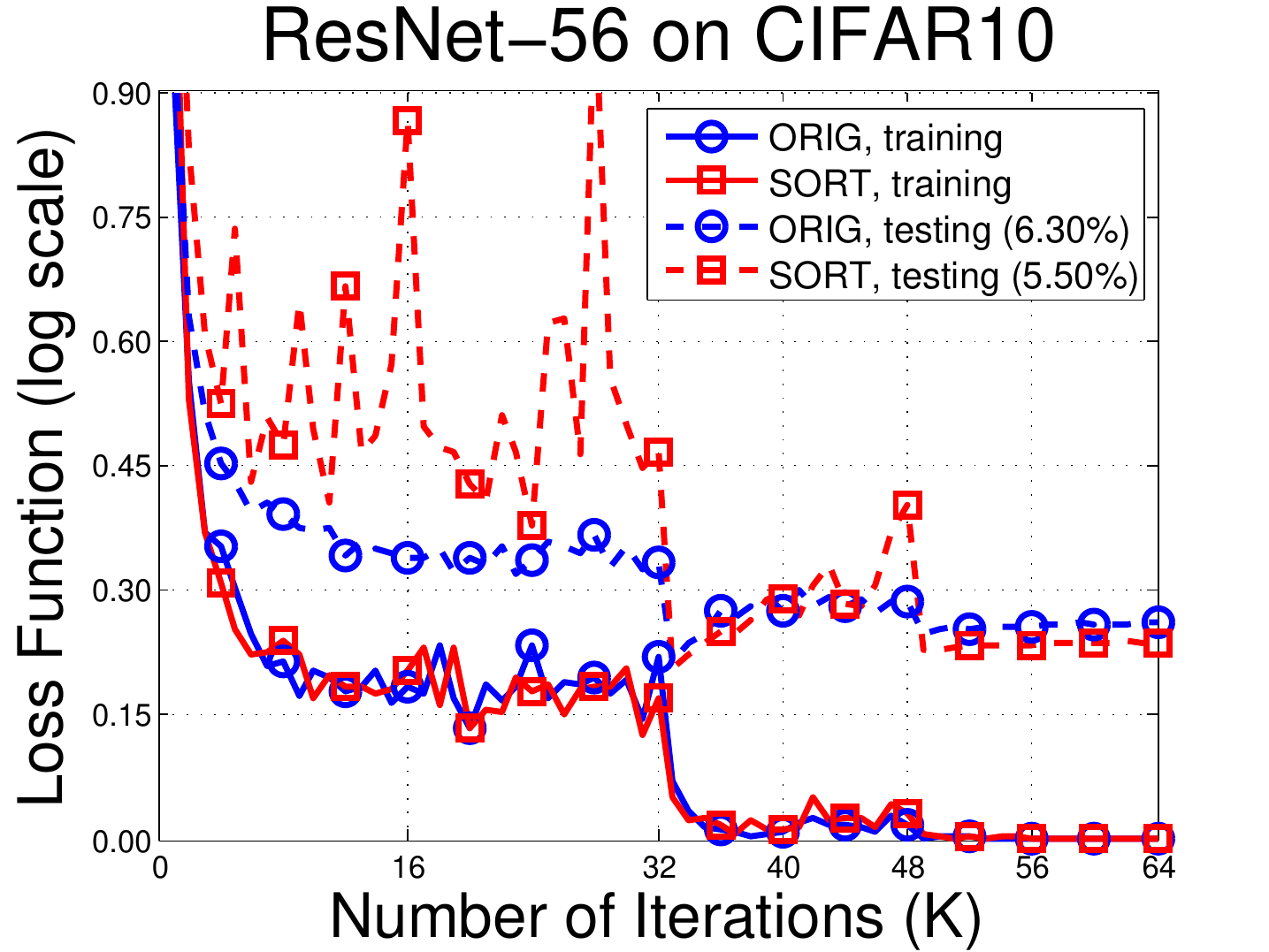}
\end{minipage}
\hspace{\subfigurehkern}
\begin{minipage}{\scatterwidth}
\centering
    \includegraphics[width=\scatterwidth]{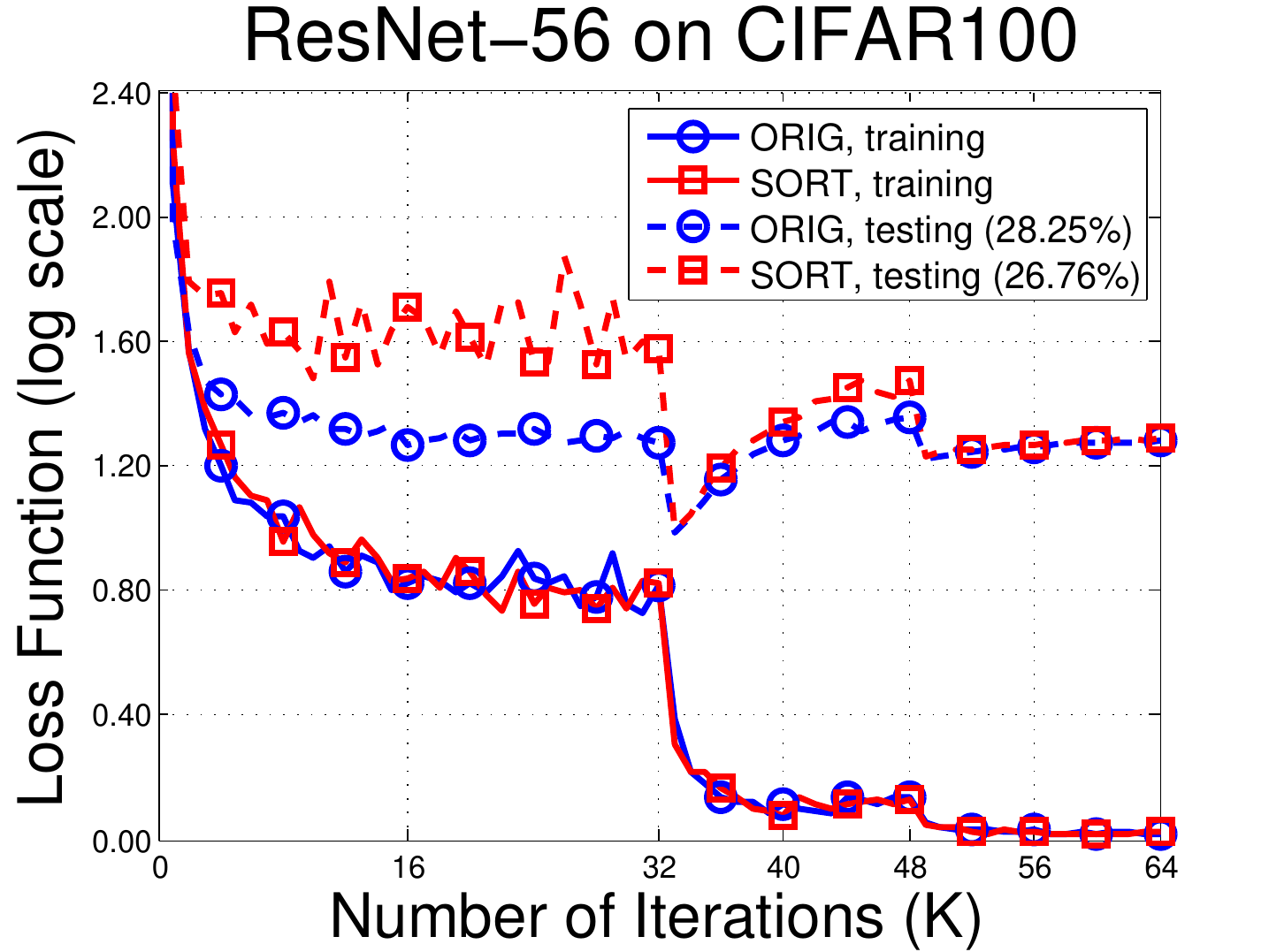}
\end{minipage}
\hspace{\subfigurehkern}
\begin{minipage}{\scatterwidth}
\centering
    \includegraphics[width=\scatterwidth]{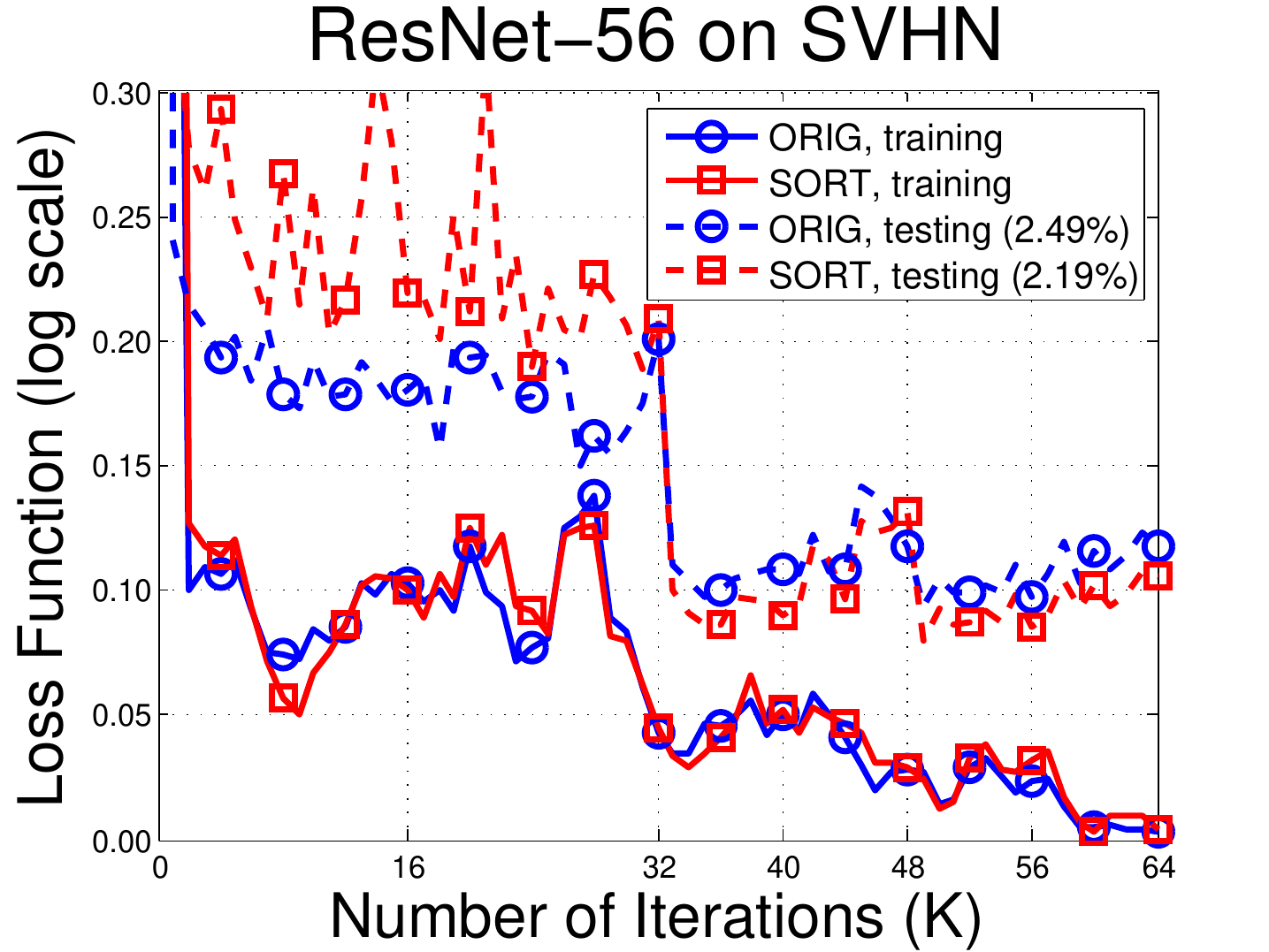}
\end{minipage} \\
\vspace{\subfigurevkern}
\begin{minipage}{\scatterwidth}
\centering
    \includegraphics[width=\scatterwidth]{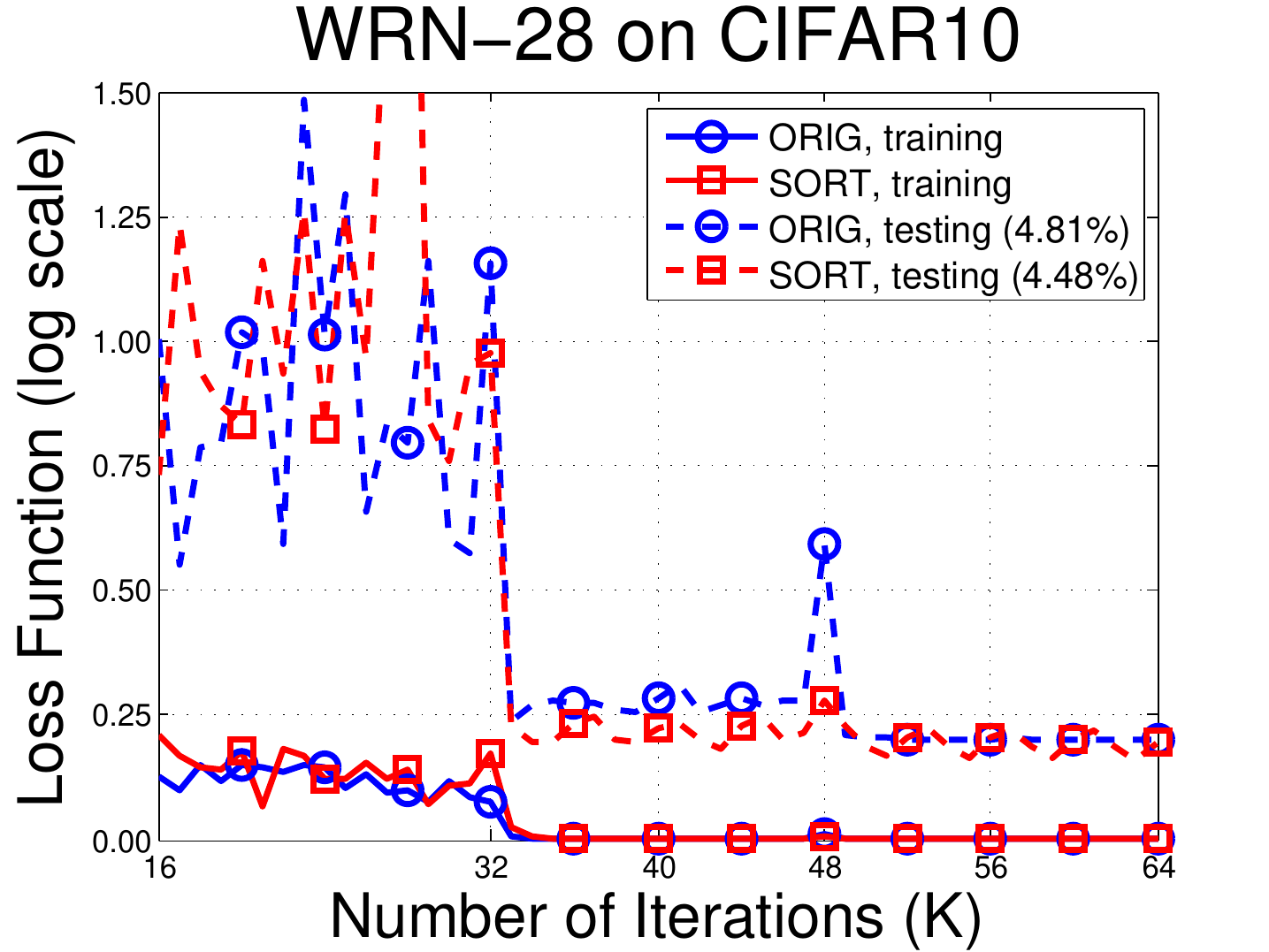}
\end{minipage}
\hspace{\subfigurehkern}
\begin{minipage}{\scatterwidth}
\centering
    \includegraphics[width=\scatterwidth]{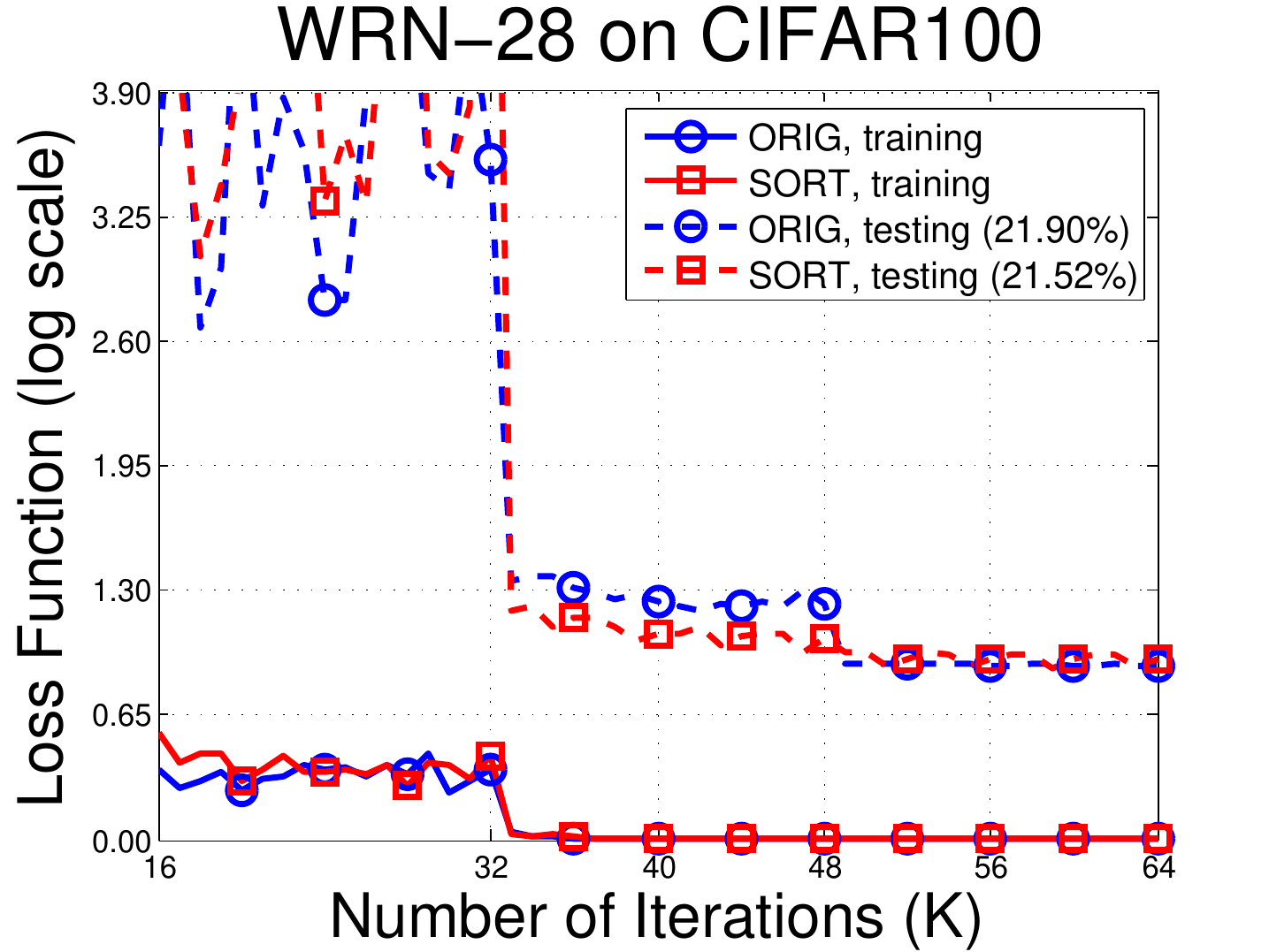}
\end{minipage}
\hspace{\subfigurehkern}
\begin{minipage}{\scatterwidth}
\centering
    \includegraphics[width=\scatterwidth]{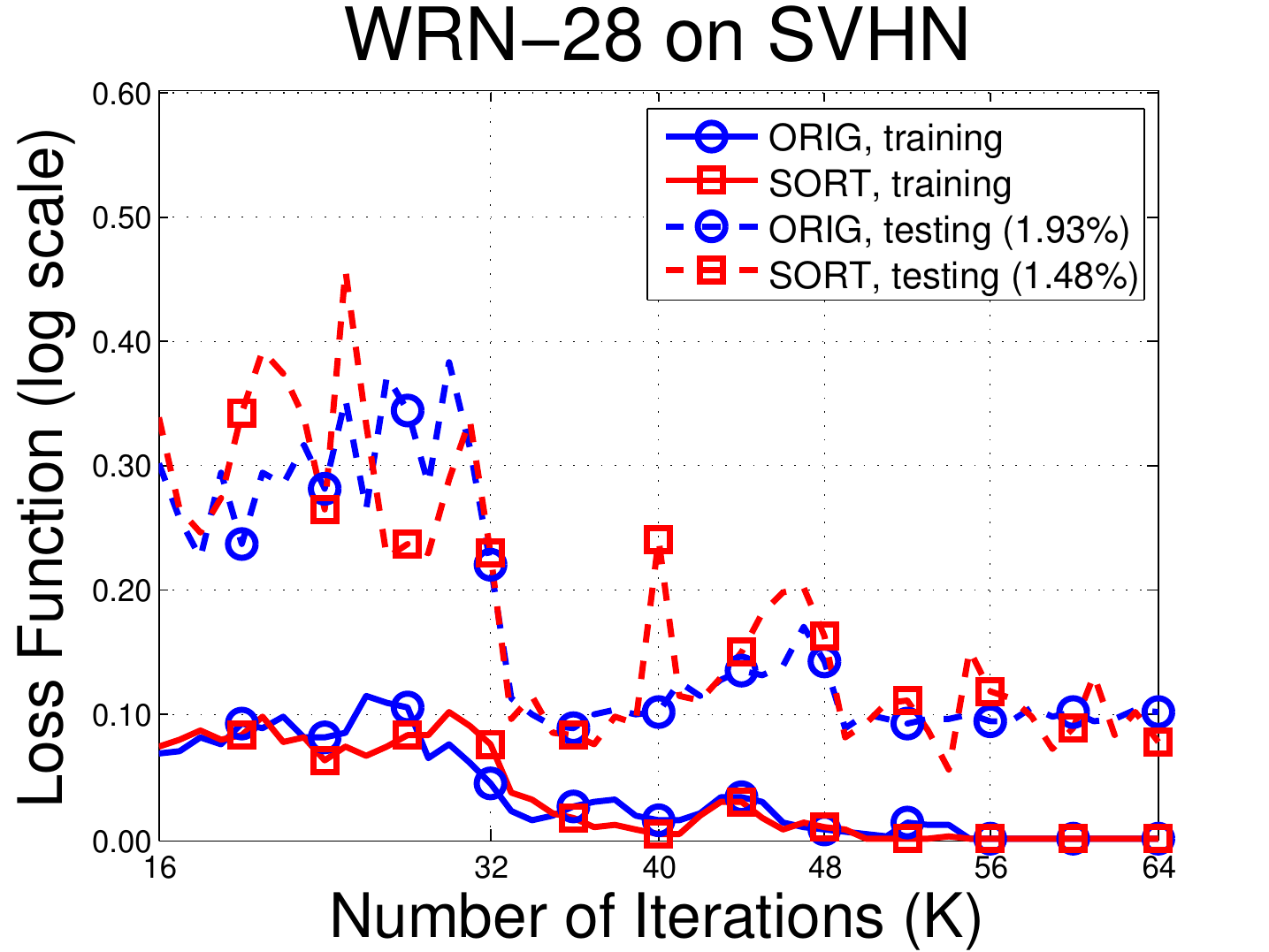}
\end{minipage}
\end{center}
\caption{
    {\bf CIFAR10}, {\bf CIFAR100} and {\bf SVHN} learning curves with different networks.
    Each number in parentheses denote the recognition error rate reported by the final model.
    Please zoom in for more details.
}
\label{Fig:SmallDatasetCurves}
\end{figure*}

We plot the learning curves of several architectures in Figure~\ref{Fig:SmallDatasetCurves}.
It is interesting to observe the convergence of network structures before and after using SORT.
On the two-branch variants of both {\bf LeNet} and {\bf BigNet},
SORT allows each parameterized branch to update its weights based on the information of the other one,
therefore it helps the network to get trained better (the testing curves are closer to $0$).
On the residual networks, as explained in Section~\ref{Approach:Nonlinearity},
SORT introduces numerical instability and makes it more difficult for the network training to converge,
thus in the first training section ({\em i.e.}, with the largest learning rate),
the network with SORT often reports unstable loss values and recognition rates compared to the network without SORT.
However, in the later sections, as the learning rate goes down and the training process becomes stable,
the network with SORT benefits from the increasing representation ability and thus works better than the baseline.
In addition, a comparable loss value of SORT can lead to better recognition accuracy
(see the curves of {\bf ResNet-56} and {\bf WRN-28} on {\bf CIFAR100}).

\subsection{ImageNet Experiments}
\label{Experiments:ImageNet}

\renewcommand{\scatterwidthA}{5.6cm}
\renewcommand{\scatterwidthB}{2.916cm}
\renewcommand{\subfigurehkern}{0.2cm}
\begin{figure*}
\begin{center}
\begin{minipage}{\scatterwidthA}
\centering
    \includegraphics[width=\scatterwidthA]{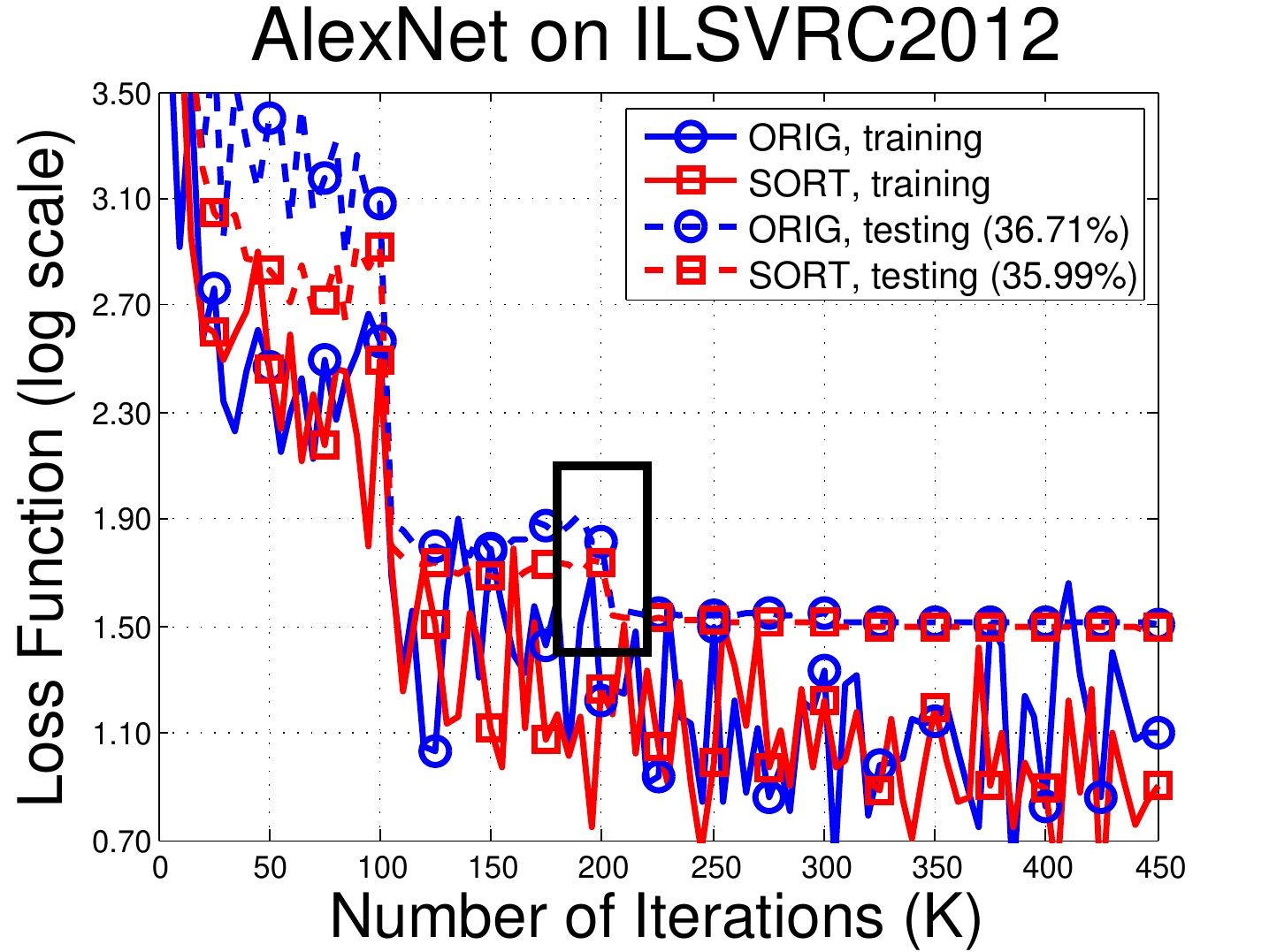}
\end{minipage}
\hspace{-0.5cm}
\begin{minipage}{\scatterwidthB}
\centering
    \includegraphics[width=\scatterwidthB]{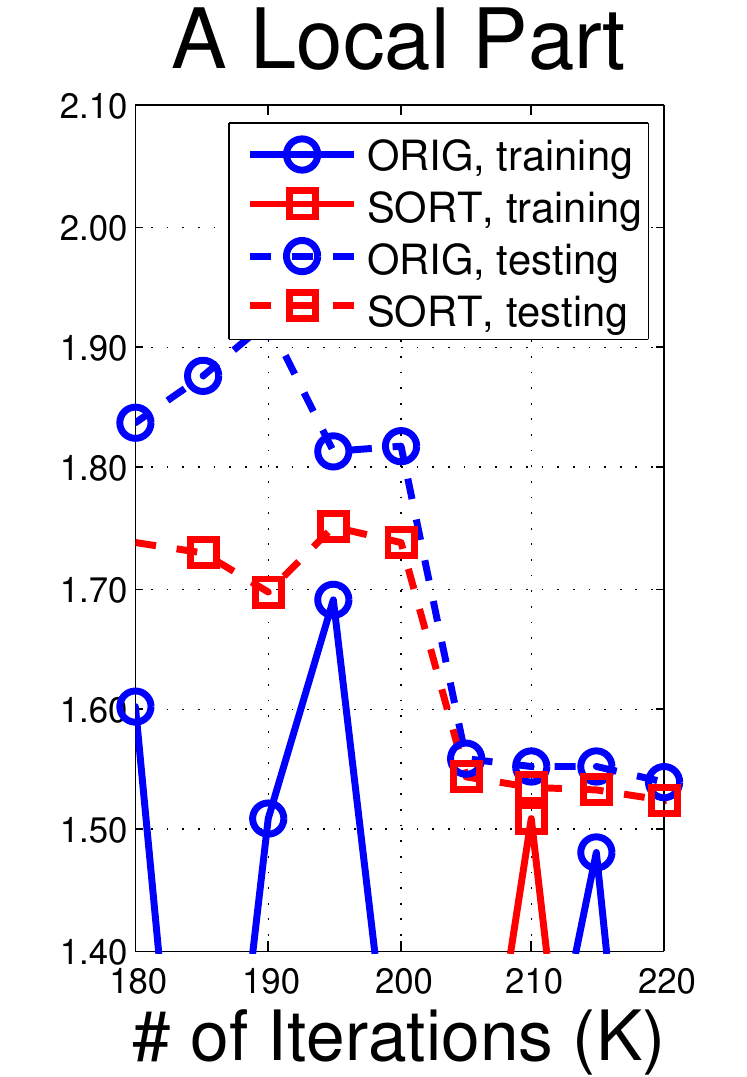}
\end{minipage}
\hspace{\subfigurehkern}
\begin{minipage}{\scatterwidthA}
\centering
    \includegraphics[width=\scatterwidthA]{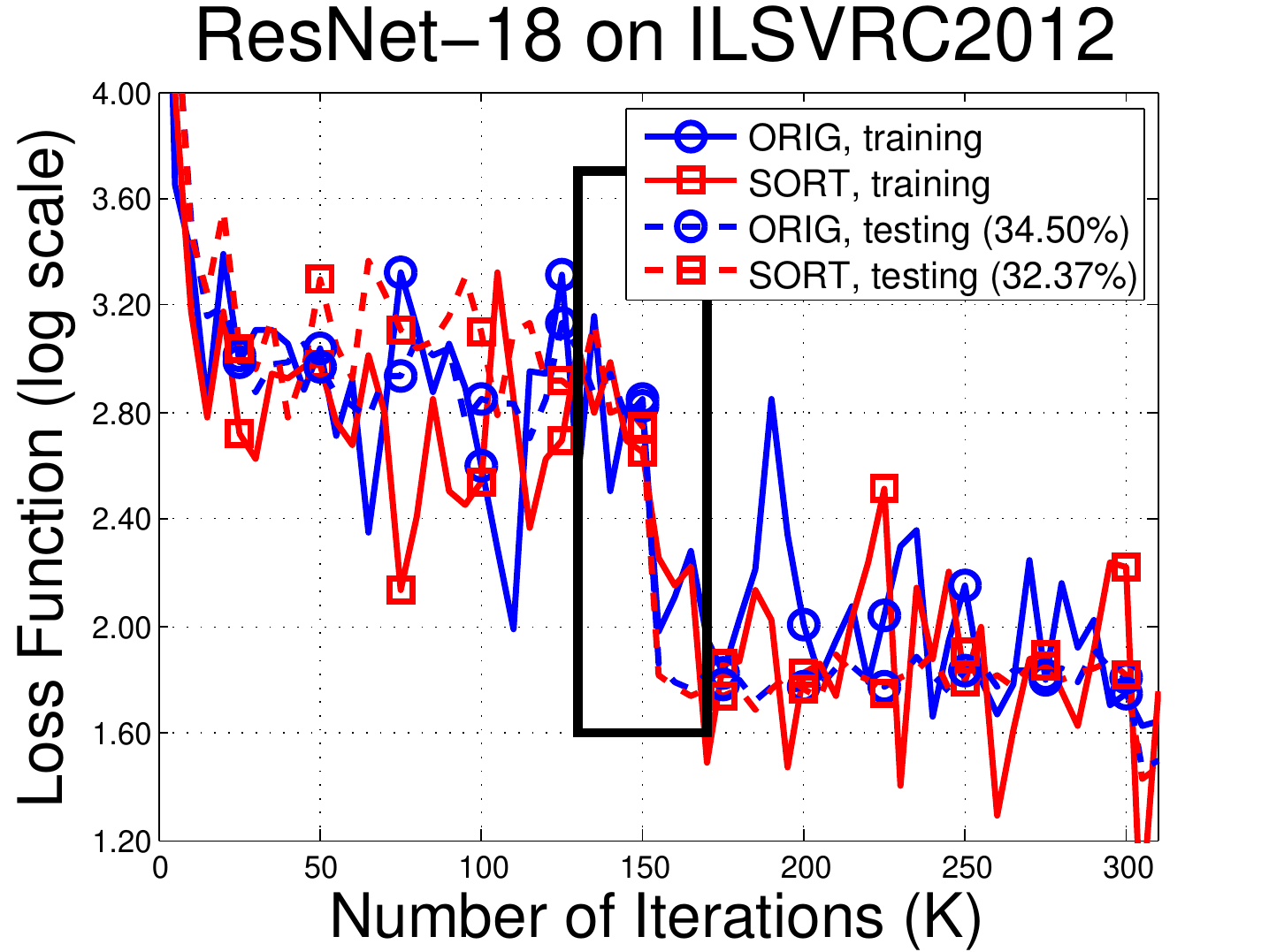}
\end{minipage}
\hspace{-0.5cm}
\begin{minipage}{\scatterwidthB}
\centering
    \includegraphics[width=\scatterwidthB]{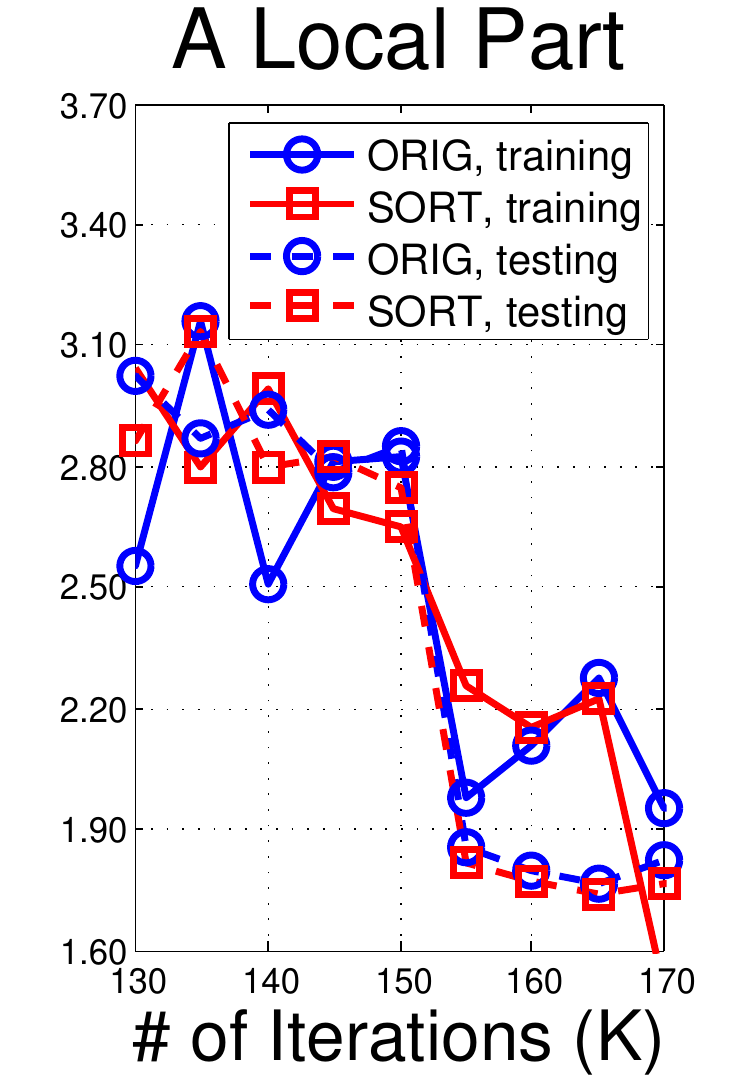}
\end{minipage}
\end{center}
\caption{
    {\bf ILSVRC2012} learning curves with {\bf AlexNet} (left) and {\bf ResNet-18} (right).
    Each number in parentheses denotes the top-$1$ error rate reported by the final model.
    For better visualization, we zoom in on a local part (marked by a black rectangle) of each learning curve.
}
\label{Fig:ImageNetCurves}
\end{figure*}

\subsubsection{Settings}
\label{Experiments:ImageNet:Settings}

We further evaluate our approach on the {\bf ILSVRC2012} dataset~\cite{Russakovsky_2015_ImageNet}.
This is a subset of the {\bf ImageNet} database~\cite{Deng_2009_ImageNet} which contains $1\rm{,}000$ object categories.
We train our models on the training set containing $1.3\mathrm{M}$ images,
and test them on the validation set containing $50\mathrm{K}$ images.
Two network architectures are taken as the baseline.
The first one is the {\bf AlexNet}~\cite{Krizhevsky_2012_ImageNet},
a $8$-layer network which is used for testing chain-styled architectures.
As in the previous experiments, we replace each of the $5$ convolutional kernels with a two-branch module,
leading to a deeper and more powerful network structure, which is denoted as {\bf AlexNet*}.
The second baseline is {\bf ResNet}~\cite{He_2016_Deep} with different numbers of layers,
which is the state-of-the-art network architecture for this large-scale visual recognition task.
In both cases, we start from scratch, and train the networks with mini-batches of $256$ images.
The {\bf AlexNet} is trained through $450\mathrm{K}$ iterations,
and the learning rate starts from $0.1$ and drops by $1/10$ after each $100\mathrm{K}$ iterations.
These numbers are $600\mathrm{K}$, $0.1$ and $150\mathrm{K}$, respectively, for training a {\bf ResNet}.

\subsubsection{Results}
\label{Experiments:ImageNet:Results}

\renewcommand{\colwidth}{1.8cm}
\begin{table}
\centering
\begin{tabular}{|l||R{\colwidth}|R{\colwidth}|}
\hline
Network               & Top-$1$ Error    & Top-$5$ Error    \\
\hline\hline
{\bf AlexNet}         & $43.19$          & $19.87$          \\
\hline
{\bf AlexNet*}        & $36.71$          & $14.77$          \\
\hline
{\bf AlexNet*}-SORT   & $\mathbf{35.99}$ & $\mathbf{14.46}$ \\
\hline\hline
{\bf ResNet-18}       & $34.50$          & $13.33$          \\
\hline
{\bf ResNet-18}-SORT  & $\mathbf{32.37}$ & $\mathbf{12.61}$ \\
\hline\hline
{\bf ResNetT-18}      & $30.50$          & $11.07$          \\
\hline
{\bf ResNetT-18}-SORT & $\mathbf{29.95}$ & $\mathbf{10.80}$ \\
\hline
{\bf ResNetT-34}      & $27.02$          & $ 8.77$          \\
\hline
{\bf ResNetT-34}-SORT & $\mathbf{26.57}$ & $\mathbf{ 8.55}$ \\
\hline
{\bf ResNetT-50}      & $24.10$          & $ 7.11$          \\
\hline
{\bf ResNetT-50}-SORT & $\mathbf{23.82}$ & $\mathbf{ 6.72}$ \\
\hline
\end{tabular}
\caption{
    Recognition error rate ($\%$) on the {\bf ILSVRC2012} dataset using different network architectures.
    All the results are reported using {\bf one single crop} in testing.
    The {\bf ResNet-18} is implemented with CAFFE, while {\bf ResNetT}'s are implemented with Torch~\cite{Gross_2016_ResNet}.
}
\label{Tab:ILSVRC2012}
\end{table}

The recognition results are summarized in Table~\ref{Tab:ILSVRC2012}.
All the numbers are reported by one single model.
Based on the original chain-styled {\bf AlexNet},
replacing each convolutional layer as a two-branch module produces $36.71\%$ top-$1$ and $14.77\%$ top-$5$ error rates,
which is significantly lower than the original version, {\em i.e.}, $43.19\%$ and $19.87\%$.
This is mainly due to the increase in network depth.
SORT further reduces the errors by $0.72\%$ and $0.31$ (or $1.96\%$ and $2.10\%$ relatively).
On the $18$-layer {\bf ResNet}, the baseline top-$1$ and top-$5$ error rates are $34.50\%$ and $13.33\%$,
and SORT reduces them to $32.37\%$ and $12.61\%$ ($6.17\%$ and $5.71\%$ relative drop, respectively).

On a $4$-GPU machine,
{\bf AlexNet*} and {\bf ResNet-18} need an average of $10.5\mathrm{s}$ and $19.3\mathrm{s}$ to finish $20$ iterations.
After SORT is applied, these numbers becomes $10.7\mathrm{s}$ and $19.9\mathrm{s}$, respectively.
Given that only less than $5\%$ extra time and no extra memory are used,
we can claim the effectiveness and the efficiency of SORT in large-scale visual recognition.

\subsubsection{Discussions}
\label{Experiments:ImageNet:Discussions}

We also plot the learning curves of both architectures in Figure~\ref{Fig:ImageNetCurves}.
Very similar phenomena are observed as in small-scale experiments.
On {\bf AlexNet*} which is the branched version of a chain-styled network, SORT helps the network to be trained better.
Meanwhile, on {\bf ResNet-18}, SORT makes the network more difficult to converge.
But nevertheless, in either cases,
SORT improves the representation ability and eventually helps the modified structure achieve better recognition performance.

\subsection{Transfer Learning Experiments}
\label{Experiments:Transfer}

\renewcommand{\colwidth}{1.00cm}
\begin{table}
\centering
\begin{tabular}{|l||R{\colwidth}|R{\colwidth}|R{\colwidth}|}
\hline
Network              & {\em pool-5}     & {\em fc-6}       & {\em fc-7}       \\
\hline\hline
{\bf AlexNet}        & $69.19$          & $71.51$          & $69.47$          \\
(std deviation)      & $\pm0.18$        & $\pm0.25$        & $\pm0.11$        \\
\hline
{\bf AlexNet*}       & $74.20$          & $76.54$          & $74.42$          \\
(std deviation)      & $\pm0.17$        & $\pm0.30$        & $\pm0.18$        \\
\hline
{\bf AlexNet*}-SORT  & $\mathbf{74.88}$ & $\mathbf{77.12}$ & $\mathbf{75.06}$ \\
(std deviation)      & $\pm0.19$        & $\pm0.24$        & $\pm0.15$        \\
\hline
\end{tabular}
\caption{
    Classification accuracy ($\%$) on the {\bf Caltech256} dataset
    using deep features extracted from different layers of different network structures.
}
\label{Tab:Classification}
\end{table}

We evaluate the transfer ability of the trained models by applying them to other image classification tasks.
The {\bf Caltech256}~\cite{Griffin_2007_Caltech} dataset is used for generic image classification.
We use the {\bf AlexNet}-based models to extract from the {\em pool-5}, {\em fc-6} and {\em fc-7} layers,
and adopt ReLU activation to filter out negative responses.
The neural responses from the {\em pool-5} layer ($6\times6\times256$) are spatially averaged into a $256$-dimensional vector,
while the other two layers directly produce $4\rm{,}096$-dimensional feature vectors.
We perform square-root normalization followed by $\ell_2$ normalization,
and use {\bf LIBLINEAR}~\cite{Fan_2008_LIBLINEAR} as an SVM implementation and set the slacking variable ${C}={10}$.
$60$ images per category are left out for training the SVM model, and the remaining ones are used for testing.
The average accuracy over all categories is reported.
We run $10$ individual training/testing splits and report the averaged accuracy as well as the standard deviation.
Results are summarized in Table~\ref{Tab:Classification}.
One can observe that the improvement on {\bf ILSVRC2012} brought by SORT is able to transfer to {\bf Caltech256}.

\section{Conclusions}
\label{Conclusions}

In this paper, we propose Second-Order Response Transform ({\bf SORT}),
an extremely simple yet effective approach to improve the representation ability of deep neural networks.
SORT summarizes two neural responses by considering both sum and product terms,
which leads to efficient information propagation throughout the network and more powerful network nonlinearity.
SORT can be applied to a wide range of modern convolutional neural networks,
and produce consistent recognition accuracy gain on some popular benchmarks.
We also verify the increasing effectiveness of SORT on very deep networks.

In the future, we will investigate the extension of SORT.
It remains open problems that whether SORT can be applied to multi-branch networks
such as Inception~\cite{Szegedy_2015_Going}, DenseNet~\cite{Huang_2017_Densely} and ResNeXt~\cite{Xie_2017_Aggregated},
or some other applications such as GANs~\cite{Goodfellow_2014_Generative} or LSTMs~\cite{Hochreiter_1997_Long}.

\vspace{0.2cm}
\noindent
{\bf Acknowledgements.}
This work was supported by the High Tech Research and Development Program of China 2015AA015801, NSFC 61521062, STCSM 12DZ2272600,
the IARPA via DoI/IBC contract number D16PC00007, and ONR N00014-15-1-2356.
We thank Xiang Xiang and Zhuotun Zhu for instructive discussions.

{\small
\bibliographystyle{ieee}
\bibliography{egbib}
}

\end{document}